# Unsupervised Methods for Determining Object and Relation Synonyms on the Web

**Alexander Yates**                                    YATES@TEMPLE.EDU
*Temple University*
*Computer and Information Sciences*
*1805 N. Broad St.*
*Wachman Hall 303A*
*Philadelphia, PA 19122*

**Oren Etzioni**                                ETZIONI@CS.WASHINGTON.EDU
*University of Washington*
*Computer Science and Engineering*
*Box 352350*
*Seattle, WA 98195-2350*

## Abstract

The task of identifying synonymous relations and objects, or *synonym resolution*, is critical for high-quality information extraction. This paper investigates synonym resolution in the context of unsupervised information extraction, where neither hand-tagged training examples nor domain knowledge is available. The paper presents a scalable, fully-implemented system that runs in $O(KN \log N)$ time in the number of extractions, $N$, and the maximum number of synonyms per word, $K$. The system, called Resolver, introduces a probabilistic relational model for predicting whether two strings are co-referential based on the similarity of the assertions containing them. On a set of two million assertions extracted from the Web, Resolver resolves objects with 78% precision and 68% recall, and resolves relations with 90% precision and 35% recall. Several variations of Resolver's probabilistic model are explored, and experiments demonstrate that under appropriate conditions these variations can improve F1 by 5%. An extension to the basic Resolver system allows it to handle polysemous names with 97% precision and 95% recall on a data set from the TREC corpus.

## 1. Introduction

Web Information Extraction (WIE) systems (Zhu, Nie, Wen, Zhang, & Ma, 2005; Agichtein, 2006; Etzioni, Cafarella, Downey, Kok, Popescu, Shaked, Soderland, Weld, & Yates, 2005) extract *assertions* that describe a relation and its arguments from Web text. For example:

<div align="center">

`(is capital of, D.C., United States)`

</div>

WIE systems can extract hundreds of millions of assertions containing millions of different strings from the Web (e.g., the TextRunner system by Banko, Cafarella, Soderland, Broadhead, & Etzioni, 2007). One problem that becomes a real challenge at this scale is that WIE systems often extract assertions that describe the same real-world object or relation using different names. For example, a WIE system might also extract

```
(is capital city of, Washington, U.S.)
```

which describes the same relationship as above but contains a different name for the relation and each argument.

Synonyms are prevalent in text, and the Web corpus is no exception. Our data set of two million assertions extracted from a Web crawl contained over a half-dozen different names each for the `United States` and `Washington, D.C.`, and three for the `is capital of` relation. The top 80 most commonly extracted objects had an average of 2.9 extracted names per entity, and several had as many as 10 names. The top 100 most commonly extracted relations had an average of 4.9 synonyms per relation.

We refer to the problem of identifying synonymous object and relation names as *synonym resolution*. Previous techniques have focused on one particular aspect of the problem, either objects or relations. In addition, these techniques often depend on a large set of training examples, or are tailored to a specific domain by assuming knowledge of the domain's schema. Due to the number and diversity of the relations extracted, these techniques are not feasible for WIE systems. Schemata are not available for the Web, and hand-labeling training examples for each relation would require a prohibitive manual effort.

In response, we present RESOLVER, a novel, domain-independent, unsupervised synonym resolution system that applies to both objects and relations. RESOLVER clusters synonymous names together using a probabilistic model informed by string similarity and the similarity of the assertions containing the names. Its similarity metric outperforms those used by similar systems for cross-document entity coreference (e.g., Mann & Yarowsky, 2003) and paraphrase discovery (Lin & Pantel, 2001; Hasegawa, Sekine, & Grishman, 2004) on their respective tasks of object and relation synonym resolution. The key questions answered by RESOLVER include:

1. *Is it possible to effectively cluster strings in a large set of extractions into sets of synonyms without using domain knowledge, manually labeled training data, or other external resources that are unavailable in the context of Web Information Extraction?* Experiments below include an empirical demonstration that RESOLVER can resolve objects with 78% precision and 68% recall, and relations with 90% precision and 35% recall.

2. *How can we scale synonym resolution to large, high-dimensional data sets?* RESOLVER provides a scalable clustering algorithm that runs in time $O(KN \log N)$ in the number of extractions, $N$, and the maximum number of synonyms per word, $K$. In theory it compares well with even fast approximate solutions for clustering large data sets in large-dimensional spaces, and in practice RESOLVER has been successfully run on a set of assertions extracted from over 100 million Web pages.

3. *How can we formalize unsupervised synonym resolution, and is there a practical benefit to doing so?* RESOLVER provides an unsupervised, generative probabilistic model for predicting whether two object or relation names co-refer, and experiments show that this significantly outperforms previous metrics for distributional similarity. In particular, it outperforms a related metric based on mutual information (Lin & Pantel, 2001) by 193% in AUC on object clustering, and by 121% on relation clustering.

4. *Is it possible to use the special properties of functions and inverse functions to improve the precision of a synonym resolution algorithm?* The basic version of Resolver's probabilistic model for object synonymy is independent of the relation in the extraction. However, it is intuitively clear that certain relations, especially functions and inverse functions, provide especially strong evidence for and against synonymy. Several extensions to the Resolver system show that without hurting recall, the precision of object merging can be improved by 3% using functions.

5. *Can Resolver handle polysemous names, which have different meanings in different contexts?* While the basic version of Resolver assumes that every name has a single meaning, we present an extension to the basic system that is able to automatically handle polysemous names. On a manually-cleaned data set of polysemous named entities from the TREC corpus, Resolver achieves a precision of 97.3% and a recall of 94.7% in detecting proper noun coreference relationships, and is able to outperform previous work in accuracy while requiring only a large, unannotated corpus as input.

The next section discusses previous work in synonym resolution. Section 3 describes the problem of synonym resolution formally and introduces notation and terminology that will be used throughout. Section 4 introduces Resolver's probabilistic model. Section 5 describes Resolver's clustering algorithm. Section 6 presents experiments with the basic Resolver system that compare its performance with the performance of previous work in synonym resolution. Section 7 describes several extensions to the basic Resolver system, together with experiments illustrating the gains in precision and recall. Section 8 develops an extension to Resolver that relaxes the assumption that every string has a single referent, and it compares Resolver experimentally to previous work in cross-document entity resolution. Finally, Section 9 discusses conclusions and areas for future work.

## 2. Previous Work

Synonym resolution encompasses two tasks, finding synonyms for extracted objects and relations. Synonym resolution for objects is very similar to the task of cross-document entity resolution (Bagga & Baldwin, 1998), in which the objective is to cluster occurrences of named entities from multiple documents into coreferential groups. Pedersen and Kulkarni (Pedersen & Kulkarni, 2007; Kulkarni & Pedersen, 2008) cluster people's names in Web documents and in emails using agglomerative clustering and a heuristic similarity function. Li, Morie, and Roth (2004a, 2004b) use an Expectation-Maximization with a graphical model and databases of common nicknames, honorifics, titles, *etc.* to achieve high accuracy on a cross-document entity resolution task. Mann and Yarowsky (2003) use a combination of extracted features and term vectors including proper names in context to cluster ambiguous names on the Web. They use the Cosine Similarity Metric (Salton & McGill, 1983) together with hierarchical agglomerative clustering. Resolver's main contribution to this body of work is that it proposes a new, formal similarity measure that works for both objects and relations, and it demonstrates both theoretically and empirically that it can scale up to millions of extractions. The Web People Search Task (WEPS) (Artile, Sekine, & Gonzalo, 2008), part of SemEval 2007, involved 16 systems trying to determine clusters of documents containing references to the same entity for ambiguous person names like "Kennedy." In Section 6, we show that RESOLVER significantly outperforms the Cosine Similarity Metric in clustering experiments. Further experiments below (Section 8) show that RESOLVER is able to achieve similar, slightly higher performance than Li *et al.* on their dataset, while not relying on any resources besides a large corpus.

Coreference resolution systems, like synonym resolution systems, try to merge references to the same object, and they apply to arbitrary noun phrases rather than just to named entities. Because of the difficulty of this general problem, most work has considered techniques informed by parsers (e.g., Lappin & Leass, 1994) or training data (e.g., Ng & Cardie, 2002; McCarthy & Lehnert, 1995). Cardie and Wagstaff (1999) use a set of extracted grammatical and semantic features and an ad-hoc clustering algorithm to perform unsupervised coreference resolution, achieving better performance on the MUC-6 coreference task than a supervised system. More recently, Haghighi and Klein (2007) use a graphical model combining local salience features and global entity features to perform unsupervised coreference, achieving an F1 score of 70.1 on MUC-6. Two systems use automatically extracted information to help make coreference resolution decisions, much like RESOLVER does. Kehler, Appelt, Taylor, and Simma (2004) use statistics over automatically-determined predicate-argument structures to compare contexts between pronouns and their potential antecedents. They find that adding this information to a system that relies on morpho-syntactic evidence for pronoun resolution provides little or no benefit. Bean and Riloff (2004) use targeted extraction patterns to find semantic constraints on the relationship between pronouns and their antecedents, and show that they can use these to improve an anaphora-resolution system. Coreference resolution is a more difficult and general task than synonym resolution for objects since it deals with arbitrary types of noun phrases. However, systems for coreference resolution also have more information available to them in the form of local sequence and salience information, which is lost in the extraction process, and they do not address relation synonymy.

Synonym resolution for relations is often called *paraphrase discovery* or *paraphrase acquisition* in NLP literature (e.g., Barzilay & Lee, 2003; Sekine, 2005). Previous work in this area (Barzilay & Lee, 2003; Barzilay & McKeown, 2001; Shinyama & Sekine, 2003; Pang, Knight, & Marcu, 2003) has looked at the use of parallel, aligned corpora, such as multiple translations of the same text or multiple news reports of the same story, to find paraphrases. Brockett and Dolan (2005) have used manually-labeled data to train a supervised model of paraphrases. The PASCAL Recognising Textual Entailment Challenge (Dagan, Glickman, & Magnini, 2006) proposes the task of recognizing when two sentences entail one another, given manually labeled training data, and many authors have submitted responses to this challenge. RESOLVER avoids the use of labor-intensive resources, and relies solely on automatically acquired extractions from a large corpus.

Several unsupervised systems for paraphrase discovery have focused on using corpus-based techniques to cluster synonymous relations. Sekine (2005) uses a heuristic similarity measure to cluster relations. Davidov and Rappoport (2008) use a heuristic clustering method to find groups of relation patterns that can be used to extract instances. Hasegawa et al. (2004) automatically extract relationships from a large corpus and cluster relations, using the Cosine Similarity Metric (Salton & McGill, 1983) and a hierarchical clustering technique like RESOLVER's. The DIRT system (Lin & Pantel, 2001) uses a similarity mea-sure based on mutual information statistics to identify relations that are similar to a given one. RESOLVER provides a formal probabilistic model for its similarity technique, and it applies to both objects and relations. Section 4.3 contains a fuller description of the differences between RESOLVER and DIRT, and Section 6 describes experiments which show RESOLVER's superior performance in precision and recall over clustering using the mutual information similarity metric employed by DIRT, as well as the Cosine Similarity Metric.

RESOLVER's method of determining the similarity between two strings is an example of a broad class of metrics called *distributional similarity* metrics (Lee, 1999), but it has significant advantages over traditional distributional similarity metrics for the synonym resolution task. All of these metrics are based on the underlying assumption, called the Distributional Hypothesis, that "Similar objects appear in similar contexts." (Hindle, 1990) Previous distributional similarity metrics, however, have been designed for comparing words based on terms appearing in the same document, rather than extracted properties. This has two important consequences: first, extracted properties are by nature sparser because they appear only in a narrow window around words and because they consist of longer strings (at the very least, pairs of words); second, each extracted shared property provides stronger evidence for synonymy than an arbitrary word that appears together with each synonym, because the extraction mechanism is designed to find meaningful relationships. RESOLVER's metric is designed to take advantage of the relational model provided by Web Information Extraction. Section 4.3 more fully describes the difference between RESOLVER's metric and the Cosine Similarity Metric (Salton & McGill, 1983), an example of a traditional distributional similarity metric. Experiments in Section 6 demonstrate that RESOLVER outperforms the Cosine Similarity Metric.

There are many unsupervised approaches for object resolution in databases, but unlike our algorithm these approaches depend on a known, fixed, and generally small schema. Ravikumar and Cohen (2004) present an unsupervised approach to object resolution using Expectation-Maximization on a hierarchical graphical model. Several other recent approaches leverage domain-specific information and heuristics for object resolution. For example, many (Dong, Halevy, & Madhavan, 2005; Bhattacharya & Getoor, 2005, 2006) rely on evidence from observing which strings appear as arguments to the same relation simultaneously (e.g., co-authors of the same publication). While this is useful information when resolving authors in the citation domain, it is rare to find relations with similar properties in extracted assertions. None of these approaches applies to the problem of resolving relations. Winkler (1999) provides a survey of this area. Several supervised learning techniques make entity resolution decisions (Kehler, 1997; McCallum & Wellner, 2004; Singla & Domingos, 2006), but of course these systems depend on the availability of training data, and even on a significant number of labeled examples per relation of interest.

One promising new approach to clustering in a relational domain is the Multiple Relational Clusterings (MRC) algorithm (Kok & Domingos, 2007). This approach, though not specific to synonym resolution, can find synonyms in a set of unlabeled, relational extractions without domain-specific heuristics. The approach is quite recent, and so far no detailed experimental comparison has been conducted.

RESOLVER's probabilistic model is partly inspired by the ball-and-urns abstraction of information extraction presented by Downey, Etzioni, and Soderland (2005) RESOLVER's task and probability model are different from theirs, but many of the same modeling as-sumptions (such as the independence of extractions) are made in both cases to simplify the derivation of the models.

Previous work on RESOLVER (Yates & Etzioni, 2007) discussed the basic version of the probabilistic model and initial experimental results. This work expands on the previous work in that it includes a new experimental comparison with an established mutual information-based similarity metric; a new extension to the basic system (property weighting); full proofs for three claims; and a description of a fast algorithm for calculating the Extracted Shared Property model.

## 3. The Formal Synonym Resolution Problem

A synonym resolution system for WIE takes a set of extractions as input and returns a set of clusters, with each cluster containing synonymous object strings or relation strings. More precisely, the input is a data set $D$ containing extracted *assertions* of the form $a = (r, o_1, \ldots, o_n)$, where $r$ is a relation string and each $o_i$ is an object string representing the arguments to the relation. Throughout this work, all assertions are assumed to be binary, so $n = 2$.

The output of a synonym resolution system is a *clustering*, or set of clusters, of the strings in $D$. Let $S$ be the set of all distinct strings in $D$. A clustering of $S$ is a set $C \subset 2^S$ such that all the clusters in $C$ are distinct, and they cover the whole set:

$$\bigcup_{c \in C} = S$$

$$\forall c_1, c_2 \in C. \, c_1 \cap c_2 = \emptyset$$

Each cluster in the output clustering constitutes the system's conjecture that all strings inside the cluster are synonyms, and no string outside that cluster is a synonym of any string in the cluster.

### 3.1 The Single-Sense Assumption

The formal representation of synonym resolution described above makes an important simplifying assumption: it is assumed that every string belongs to exactly one cluster. In language, however, strings often have multiple meanings; *i.e.*, they are *polysemous*. Polysemous strings cannot be adequately represented using a clustering in which each string belongs to exactly one cluster. For most of this paper, we will make the single-sense assumption, but Section 8 illustrates an extension to RESOLVER that does away with this assumption.

As an example of the representational trouble posed by polysemy, consider the name "President Roosevelt." In certain contexts, this name is synonymous with "President Franklin D. Roosevelt," and in other contexts it is synonymous with "President Theodore Roosevelt." However, "President Franklin D. Roosevelt" is never synonymous with "President Theodore Roosevelt." There is no clustering of the three names, using the notion of clustering described above, such that all synonymy relationships are accurately represented.

Others have described alternate kinds of clustering that take polysemy into account. For example, "soft clustering" allows a string to be assigned to as many different clusters as it has senses. One variation on this idea is to assign a probability distribution to every string, describing the prior probability that the string belongs to each cluster (Li & Abe, 1998; Pereira, Tishby, & Lee, 1993). Both of these representations capture only prior information about strings. That is, they represent the idea that a particular string can belong to a cluster, or the probability that it belongs to a cluster, but not whether a particular instance of the string actually does belong to a cluster. A third type of clustering, the most explicit representation, stores each instance of a string separately. Each string instance is assigned to the cluster that is most appropriate for the instance's context. Word sense disambiguation systems that assign senses from WordNet (Miller, Beckwith, Fellbaum, Gross, & Miller., 1990) implicitly use this kind of clustering (e.g., Ide & Veronis, 1998; Sinha & Mihalcea, 2007).

## 3.2 Subproblems in Synonym Resolution

The synonym resolution problem can be divided into two subproblems: first, how to measure the similarity, or probability of synonymy, between pairs of strings in $S$; and second, how to form clusters such that all of the elements in each cluster have high similarity to one another, and relatively low similarity to elements in other clusters.

RESOLVER uses a generative, probabilistic model for finding the similarity between strings. For strings $s_i$ and $s_j$, let $R_{i,j}$ be the random variable for the event that $s_i$ and $s_j$ refer to the same entity. Let $R_{i,j}^t$ denote the event that $R_{i,j}$ is true, and $R_{i,j}^f$ denote the event that it is false. Let $D_x$ denote the set of extractions in $D$ which contain string $x$. Given $D$ and $S$, the first subtask of synonym resolution is to find $P(R_{i,j}^t | D_{s_i}, D_{s_j})$ for all pairs $s_i$ and $s_j$. The second subtask takes $S$ and the probability scores for pairs of strings from $S$ as input. Its output is a clustering of $S$. Sections 4 and 5 cover RESOLVER's solutions to each subtask respectively.

## 4. Models for String Comparisons

Our probabilistic model provides a formal, rigorous method for resolving synonyms in the absence of training data. It has two sources of evidence: the similarity of the strings themselves (*i.e.*, edit distance) and the similarity of the assertions they appear in. This second source of evidence is sometimes referred to as *distributional similarity* (Hindle, 1990).

Section 4.1 presents a simple model for predicting whether a pair of strings are synonymous based on string similarity. Section 4.2 then presents a model called the Extracted Shared Property (ESP) Model for predicting whether a pair of strings co-refer based on their distributional similarity. Section 4.3 compares the ESP model with other methods for computing distributional similarity to give an intuition for how it behaves. Finally, Sections 4.4 and 4.5 present a method for combining the ESP model and the string similarity model to come up with an overall prediction for synonymy decisions between two clusters of strings.

## 4.1 String Similarity Model

Many objects appear with multiple names that are substrings, acronyms, abbreviations, or other simple variations of one another. Thus string similarity can be an important source of evidence for whether two strings co-refer (Cohen, 1998). RESOLVER's probabilistic String Similarity Model (SSM) assumes a similarity function $\text{sim}(s_1, s_2)$: $STRING \times STRING \to [0, 1]$. The model sets the probability of $s_1$ co-referring with $s_2$ to a smoothed version of the similarity:

$$P(R_{i,j}^t | \text{sim}(s_1, s_2)) = \frac{\alpha * \text{sim}(s_1, s_2) + 1}{\alpha + \beta}$$

As $\alpha$ increases, the probability estimate transitions from $1/\beta$ (at $\alpha = 0$) to the value of the similarity function (for very large $\alpha$). The particular choice of $\alpha$ and $\beta$ make little difference to RESOLVER's results, as long as they are chosen such that the resulting probability can never be one or zero. In the experiments below, $\alpha = 20$ and $\beta = 5$. The Monge-Elkan string similarity function (Monge & Elkan, 1996) is used for objects, and the Levenshtein string edit-distance function is used for relations (Cohen, Ravikumar, & Fienberg, 2003).

## 4.2 The Extracted Shared Property Model

The Extracted Shared Property Model (ESP) outputs the probability that two strings co-refer based on the similarity of the extracted assertions in which they appear. For example, if the extractions (`invented, Newton, calculus`) and (`invented, Leibniz, calculus`) both appeared in the data, then `Newton` and `Leibniz` would be judged to have similar contexts in the extracted data.

More formally, let a pair of strings $(r, s)$ be called a *property* of an object string $o$ if there is an assertion $(r, o, s) \in D$ or $(r, s, o) \in D$. A pair of strings $(s_1, s_2)$ is an *instance* of a relation string $r$ if there is an assertion $(r, s_1, s_2) \in D$. Equivalently, the property $p = (r, s)$ *applies* to $o$, and the instance $i = (s_1, s_2)$ *belongs* to $r$. The ESP model outputs the probability that two strings co-refer based on how many properties (or instances) they share.

As an example, consider the strings `Mars` and `Red Planet`, which appear in our data 659 and 26 times respectively. Out of these extracted assertions, they share four properties. For example, (`lacks, Mars, ozone layer`) and (`lacks, Red Planet, ozone layer`) both appear as assertions in our data. The ESP model determines the probability that `Mars` and `Red Planet` refer to the same entity after observing $k$, the number of properties that apply to both; $n_1$, the total number of extracted properties for `Mars`; and $n_2$, the total number of extracted properties for `Red Planet`.

ESP models the extraction of assertions as a generative process, much like the URNS model (Downey et al., 2005). For each string $s_i$, a certain number, $P_i$, of properties of the string are written on balls and placed in an urn. Extracting $n_i$ assertions that contain $s_i$ amounts to selecting a subset of size $n_i$ from these labeled balls.[1] Properties in the urn are called *potential properties* to distinguish them from extracted properties.

To model synonymy decisions, ESP uses a pair of urns, containing $P_i$ and $P_j$ balls respectively, for the two strings $s_i$ and $s_j$. Some subset of the $P_i$ balls have the exact same labels as an equal-sized subset of the $P_j$ balls. Let the size of this subset be $S_{i,j}$. Crucially, the ESP model assumes that synonymous strings share as many potential properties as possible, though only a few of the potential properties will be extracted for both. For non-synonymous strings, the set of shared potential properties is a strict subset of the potential properties of each string. Thus the central modeling choice in the ESP model is: *if $s_i$ and $s_j$ are synonymous (i.e., $R_{i,j} = R_{i,j}^t$) then the number of shared potential properties ($S_{i,j}$) is equal to the number of potential properties in the smaller urn ($\min(P_i, P_j)$), and if the two strings are not synonymous ($R_{i,j} = R_{i,j}^f$) then the number of shared potential properties is strictly less than the number of properties in the smaller urn ($S_{i,j} < \min(P_i, P_j)$).*

---

1. Unlike the URNS model, balls are drawn without replacement. The TEXTRUNNER data contains only one mention of any extraction, so drawing without replacement tends to model the data more accurately.

The ESP model makes several simplifying assumptions in order to make probability predictions. As is suggested by the ball-and-urn abstraction, it assumes that each ball for a string is equally likely to be selected from its urn. Because of data sparsity, almost all properties are very rare, so it would be difficult to get a better estimate for the prior probability of selecting a particular potential property. Second, balls are drawn from one urn independent of draws from any other urn. And finally, it assumes that without knowing the value of $k$, every value of $S_{i,j}$ is equally likely, since we have no better information.

Given these assumptions, we can derive an expression for $P(R_{i,j}^t)$. The derivation is sketched below; see Appendix A for a complete derivation. First, note that there are $\binom{P_i}{n_i}\binom{P_j}{n_j}$ total ways of extracting $n_i$ and $n_j$ assertions for $s_i$ and $s_j$. Given a particular value of $S_{i,j}$, the number of ways in which $n_i$ and $n_j$ assertions can be extracted such that they share exactly $k$ is given by

$$\text{Count}(k, n_i, n_j | P_i, P_j, S_{i,j}) = \binom{S_{i,j}}{k} \sum_{r,s \geq 0} \binom{S_{i,j}-k}{r+s}\binom{r+s}{r}\binom{P_i-S_{i,j}}{n_i-(k+r)}\binom{P_j-S_{i,j}}{n_j-(k+s)} \tag{1}$$

By our assumptions,

$$P(k|n_i, n_j, P_i, P_j, S_{i,j}) = \frac{\text{Count}(k, n_i, n_j | P_i, P_j, S_{i,j})}{\binom{P_i}{n_i}\binom{P_j}{n_j}} \tag{2}$$

Let $P_{\min} = \min(P_i, P_j)$. The result below follows from Bayes' Rule and our assumptions above:

**Proposition 1** *If two strings $s_i$ and $s_j$ have $P_i$ and $P_j$ potential properties (or instances), and they appear in extracted assertions $D_i$ and $D_j$ such that $|D_i| = n_i$ and $|D_j| = n_j$, and they share $k$ extracted properties (or instances), the probability that $s_i$ and $s_j$ co-refer is:*

$$P(R_{i,j}^t | D_i, D_j, P_i, P_j) = \frac{P(k|n_i, n_j, P_i, P_j, S_{i,j} = P_{\min})}{\displaystyle\sum_{\substack{S_{i,j} \\ k \leq S_{i,j} \leq P_{\min}}} P(k|n_i, n_j, P_i, P_j, S_{i,j})} \tag{3}$$

Substituting equation 2 into equation 3 gives us a complete expression for the probability we are looking for.

Note that the probability for $R_{i,j}^t$ depends on two hidden parameters, $P_i$ and $P_j$. Since in unsupervised synonym resolution there is no labeled data to estimate these parameters from, these parameters are tied to the number of times the respective strings $s_i$ and $s_j$ are extracted: $P_i = N \times n_i$. The discussion of experimental methods in Section 6 explains how the parameter $N$ is set.

Appendix B illustrates a technique for calculating the ESP model efficiently.

### 4.3 Comparison of ESP with Other Distributional Similarity Metrics

The Discovery of Inference Rules from Text (DIRT) (Lin & Pantel, 2001) system is the most similar previous work to RESOLVER in its goals, but DIRT's similarity metric is very different from ESP. Like ESP, DIRT operates over triples of extracted strings and produces similarity scores for relations by comparing the distributions of one relation's arguments to another's. The DIRT system, however, has its own extraction mechanism based on a dependency parser. Here we focus on the differences in the two systems' similarity metrics, and compare performance on the same set of extracted triples produced by TEXTRUNNER, since the extracted triples used by DIRT were not available to us. We refer to the mutual-information-based similarity metric employed by the DIRT system as $s_{MI}$. It is important to note that $s_{MI}$ as we describe it here is our own implementation of the similarity metric described by Lin and Pantel (2001), and is not the complete DIRT system.

We now briefly describe $s_{MI}$ as it applies to a set of extractions. $s_{MI}$ originally was applied to only relation strings, and for simplicity we describe it that way here, but it is readily generalized to a metric for computing the similarity between two argument-1 strings or two argument-2 strings. For notational convenience, let $D_{x=s}$ be the set of extractions that contain string $s$ at position $x$. For example, $D_{2=\text{Einstein}}$ would contain the extraction (`discovered, Einstein, Relativity`), but not the extraction (`talked with, Bohr, Einstein`). Similarly, let $D_{x=s_1,y=s_2}$ be the set of extractions that contain $s_1$ and $s_2$ at positions $x$ and $y$ respectively. Finally, let the projection of a set of extractions $D = \{(d_1, d_2, d_3)\}$ onto one of its dimensions $x$ be given by:

$$proj_x(D) = \{s | \exists_{d_1,d_2,d_3}.d_x = s \wedge (d_1, d_2, d_3) \in D\}$$

$s_{MI}$ uses a mutual information score to determine how much weight to give to each string in the set of extractions during its similarity computation. For a string $s$ at position $x$, the mutual information between it and a relation $r$ at position 1 is given by:

$$mi_{1,x}(r,s) = \log\left(\frac{|D_{1=r,x=s}| \times |D|}{|D_{1=r}| \times |D_{x=s}|}\right)$$

$s_{MI}$ calculates the similarity between two relations by first calculating the similarity between the sets of first arguments to the relations, and then the similarity between the sets of second arguments. Let $r_1$ and $r_2$ be two relations, and let the position of the argument being compared be $x$. The similarity function used is:

$$sim_x(r_1, r_2) = \frac{\displaystyle\sum_{a \in proj_x(D_{1=r_1}) \cap proj_x(D_{1=r_2})} mi_{1,x}(r_1, a) + mi_{1,x}(r_2, a)}{\displaystyle\sum_{a \in proj_x(D_{1=r_1})} mi_{1,x}(r_1, a) + \sum_{a \in proj_x(D_{1=r_2})} mi_{1,x}(r_2, a)}$$

The final similarity score for two relations is the geometric average of the similarity scores for each argument:

$$s_{MI}(r_1, r_2) = \sqrt{sim_2(r_1, r_2) \times sim_3(r_1, r_2)} \tag{4}$$

Applying the $s_{MI}$ metric to entities rather than relations simply requires projecting onto different dimensions of the relevant tuple sets.

The most significant difference between the $s_{MI}$ similarity metric and the ESP model is that the $s_{MI}$ metric compares the $x$ arguments from one relation to the $x$ arguments of the other, and then compares the $y$ arguments from one relation to the $y$ arguments of the other, and finally combines the scores. In contrast, ESP compares the $(x, y)$ argument pairs of one relation to the $(x, y)$ pairs of the other. While the $s_{MI}$ metric has the advantage that it is more likely to find matches between two relations in sparse data, it has the disadvantage that the matches it does find are not necessarily strong evidence for synonymy. In effect, it is capturing the intuition that synonyms have the same argument types for their domains and ranges, but it is certainly possible for non-synonyms to have similar domains and ranges. Antonyms are an obvious example. Synonyms are not defined by their domains and ranges, but rather by the mapping between them, and ESP better captures the similarity in this mapping. Experiments below (Section 6) compare the ESP as a similarity metric against $s_{MI}$, as given in Equation 4.

As previously mentioned, there is a large body of previous work on similarity metrics (e.g., Lee, 1999). We now compare ESP with one of the more popular of these metrics, the Cosine Similarity Metric (CSM), which has previously been used in synonym resolution work (Mann & Yarowsky, 2003; Hasegawa et al., 2004). Like most traditional distributional similarity metrics, CSM operates over context vectors, rather than extracted triples. However, the ESP model is very similar to CSM in this regard. For each extracted string, it in effect creates a binary vector of properties, ones representing properties that apply to the string and zeros representing those that do not. For example, the string `Einstein` would have a context vector with a one in the position for the property (`discovered, Relativity`), and a zero in the position for the property (`invented, light bulb`). Both ESP and CSM calculate similarities by comparing these vectors.

The specific metric used to compute CSM for two vectors $\vec{x}$ and $\vec{y}$ is given by:

$$
\begin{aligned}
sim_{CSM}(\vec{x}, \vec{y}) \quad &= \quad \frac{\vec{x} \cdot \vec{y}}{||\vec{x}|| \times ||\vec{y}||} \\
&= \quad \frac{\sum_i x_i y_i}{\sqrt{\sum_i x_i^2} \times \sqrt{\sum_i y_i^2}}
\end{aligned}
$$

Often, techniques like term weighting or TFIDF (Salton & McGill, 1983) are used with CSM to create vectors that are not boolean, but rather have dimensions with different weights according to how informative those dimensions are. We experimented with TFIDF-like weighting schemes, where the number of times an extraction was extracted is used as the term frequency, and the number of different strings a property applies to is used as the document frequency. However, we found that these weighting schemes had negative effects on performance, so from here on we ignore them. For two boolean vectors, CSM reduces to a simple computation on the number of shared properties $k$ and the number of extractions for each string, $n_1$ and $n_2$ respectively. It is given by:

$$
sim_{CSM-boolean}(\vec{x}, \vec{y}) = \frac{k}{\sqrt{n_1 n_2}} \tag{5}
$$

CSM determines how similar two context vectors are in each dimension, and then adds the scores up in a weighted sum. In contrast, ESP is highly non-linear in the number of shared properties. As the number of matching contexts grows, the weight for each additional matching context also grows. Figure 1 compares the behavior of ESP and CSM as the number of shared properties between two strings increases. Holding the number of extractions fixed and assuming boolean vectors for CSM, it behaves as a linear function of the number of shared properties. On the other hand, the ESP has the shape of a thresholding function: it has a very low value until a threshold point around $k = 10$, at which point its probability estimate starts increasing rapidly. The effect is that ESP has much lower similarity scores than CSM for small numbers of matching contexts, and much higher scores for larger numbers of matching contexts. The threshold at which it switches depends on $n_1$ and $n_2$, as well as $P_1$ and $P_2$, but we can show experimentally that our method for estimating $P_1$ and $P_2$, though simple, can be effective. Experiments in Section 6 compare the ESP model with CSM, as computed using Equation 5.

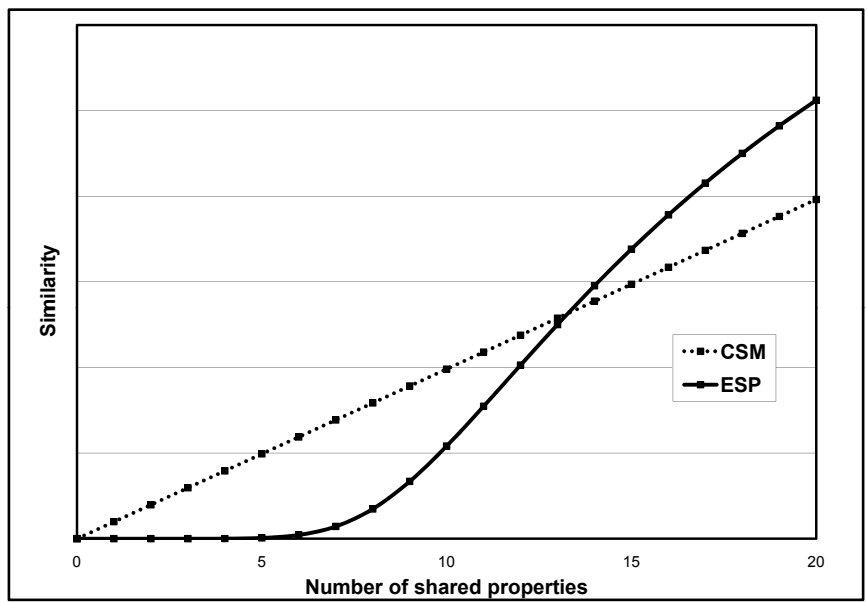

Figure 1: **The behavior of the Extracted Shared Property (ESP) model and the Cosine Similarity Model (CSM) as the number of shared properties between two strings varies.** The graph shows similarity results using two hypothetical strings with 20 extracted properties each. For ESP, the property multiple is $N = 2$. We removed scale from the $y$ axis, since the scales for the two metrics are not directly comparable, but the shape of the curves remains the same.

## 4.4 Combining the Evidence

For each potential synonymy relationship, Resolver considers two pieces of probabilistic evidence. Let $E_{i,j}^e$ be the evidence for ESP, and let $E_{i,j}^s$ be the evidence for SSM. Our method for combining the two uses the Naïve Bayes assumption that each piece of evidence is conditionally independent, given the synonymy relationship:

$$P(E^s_{i,j}, E^e_{i,j} | R_{i,j}) = P(E^s_{i,j} | R_{i,j}) P(E^e_{i,j} | R_{i,j}) \qquad (6)$$

Given this simplifying assumption, we can combine the evidence to find the probability of a coreference relationship by applying Bayes' Rule to both sides (we omit the $i, j$ indices for brevity):

$$P(R^t | E^s, E^e) = \frac{P(R^t | E^s) P(R^t | E^e)(1 - P(R^t))}{\sum_{i \in \{t, f\}} P(R^i | E^s) P(R^i | E^e)(1 - P(R^i))} \qquad (7)$$

## 4.5 Comparing Clusters of Strings

Our algorithm merges clusters of strings with one another, using the above models. However, these models give probabilities for synonymy decisions between two individual strings, not two clusters of strings.

We have experimented with several different methods of determining the probability of synonymy from the individual probability scores for each pair of strings, one taken from each cluster. Initially, we followed the work of Snow, Jurafsky, and Ng (2006) in incorporating transitive closure constraints in probabilistic modeling, and we made the same independence assumptions. This approach provides a formal probabilistic framework for the problem that is simple and efficient to calculate. In other experiments, we found that simply taking the mean or geometric mean (or even the harmonic mean) of the string pair scores provided slightly improved results. For completeness, we now provide a brief explanation of the probabilistic method for combining string pair scores into cluster pair scores.

Let a *clustering* be a set of synonymy relationships between pairs of strings such that the synonymy relationships obey the transitive closure property. We let the probability of a set of assertions $D$ given a clustering $C$ be:

$$P(D|C) = \prod_{R^t_{i,j} \in C} P(D_i \cup D_j | R^t_{i,j}) \times \prod_{R^f_{i,j} \in C} P(D_i \cup D_j | R^f_{i,j}) \qquad (8)$$

The metric used to determine if two clusters should be merged is the likelihood ratio, or the probability for the set of assertions given the merged clusters over the probability given the original clustering. Let $C'$ be a clustering that differs from $C$ only in that two clusters in $C$ have been merged in $C'$, and let $\Delta C$ be the set of synonymy relationships in $C'$ that are true, but the corresponding ones in $C$ are false. This metric is given by:

$$P(D|C')/P(D|C) = \frac{\prod_{R^t_{i,j} \in \Delta C} P(R^t_{i,j} | D_i \cup D_j)(1 - P(R^t_{i,j}))}{\prod_{R^t_{i,j} \in \Delta C} (1 - P(R^t_{i,j} | D_i \cup D_j)) P(R^t_{i,j})} \qquad (9)$$

The probability $P(R^t_{i,j} | D_i \cup D_j)$ may be supplied by SSM, ESP, or the combination model. In our experiments, we let the prior for the SSM model be 0.5. For the ESP and combined models, we set the prior to $P(R^t_{i,j}) = \frac{1}{\min(P_i, P_j)}$, where $P_i$ and $P_j$ are the number of potential properties for $s_i$ and $s_j$ respectively.

## 5. RESOLVER's Clustering Algorithm

Synonym resolution for the Web requires a clustering algorithm that can scale to a huge number of strings in a sparse, high-dimensional space. Those requirements are difficult for any clustering algorithm. On the other hand, very few words have more than a handful of synonyms, so clusters tend to be quite small. Greedy agglomerative approaches are well-suited to this type of clustering problem, since they start with the smallest possible clusters and merge them as needed.

The RESOLVER clustering algorithm is a version of greedy agglomerative clustering, with a key modification that allows it to scale to sparse, high-dimensional spaces and huge numbers of elements. A standard greedy clustering algorithm begins by comparing each pair of data points, and then greedily merges the closest pair. The biggest hurdle to scaling such an algorithm is that the initial step of comparing every pair of data points requires $O(N^2)$ comparisons for $N$ points. Several proposed techniques have been able to speed up this process in practice by filtering out some of the initial pairs of points to be compared; we build on this work to provide a novel technique with a new bound of $O(N \log N)$ comparisons, under very mild assumptions.

Our algorithm is outlined in Figure 2. It begins by calculating similarity scores between pairs of strings, in steps 1-4. Then the scores are sorted and the best cluster pairs are merged until no pair of clusters has a score above threshold. The novel part of the algorithm, step 4, compares pairs of clusters that share the same property, as long as no more than $Max$ clusters share that same property. This step limits the number of comparisons made between clusters, and it is the reason for the algorithm's improved efficiency, as explained below.

This algorithm compares every pair of clusters that have the potential to be merged, assuming two properties of the data. First, it assumes that pairs of clusters with no shared properties are not worth comparing. Since the number of shared properties is a key source of evidence for our approach, these clusters almost certainly will not be merged, even if they are compared, so the assumption is quite reasonable. Second, the approach assumes that clusters sharing only properties that apply to very many strings (at least $Max$) need not be compared. Since properties shared by many strings provide little evidence that the strings are synonymous, this assumption is reasonable for synonym resolution.

We use $Max = 50$ in our experiments. Less than 0.1% of the distinct properties are thrown out using this cutoff, but because these discarded properties apply to many strings (at least $Max$), and because the number of comparisons grows with the square of the number of strings that a property applies to, the restriction drastically cuts down on the total number of comparisons made. Table 1 shows the number of comparisons made by the naïve method of comparing all pairs of strings in a set of over 2 million extractions, and the number of comparisons that RESOLVER makes in these experiments. Our algorithm achieves a reduction by a factor of 136 for objects and 486 for relations in the number of comparisons made. An unoptimized implementation of RESOLVER is able to cluster the strings in these extractions in approximately 30 minutes. RESOLVER was also run on a larger set containing over 100 million extractions and over 1 million distinct strings, and was able to cluster these in approximately 3.5 days on a single machine.

$E := \{e = (r, a, b) | (r, a, b) \text{ is an extracted assertion}\}$
$S := \{s | s \text{ appears as a relation or argument string in } E\}$
$Cluster := \{\}$
$Elements := \{\}$
1. For each $s \in S$:
    $Cluster[s] :=$ new cluster id
    $Elements[Cluster[s]] := \{s\}$
2. $Scores := \{\}, Index := \{\}$
3. For each $e = (r, a, b) \in E$:
    $property := (a, b)$
    $Index[property] := Index[property] \cup \{Cluster[r]\}$
    $property := (r, a)$
    $Index[property] := Index[property] \cup \{Cluster[b]\}$
    $property := (r, b)$
    $Index[property] := Index[property] \cup \{Cluster[a]\}$
4. For each property $p \in Index$:
    If $|Index[p]| <$ Max:
        For each pair $\{c_1, c_2\} \subset Index[p]$:
            $Scores[\{c_1, c_2\}] :=$ similarity($c_1, c_2$)
5. Repeat until no merges can be performed:
    Sort $Scores$
    $UsedClusters := \{\}$
    Repeat until $Scores$ is empty or top score $< Threshold$:
        $\{c_1, c_2\} := removeTopPair(Scores)$
        If neither $c_1$ nor $c_2$ is in $UsedClusters$:
            $Elements[c_1] := Elements[c_1] \cup Elements[c_2]$
            For each $e \in Elements[c_2]$:
                $Cluster[e] := c_1$
            delete $c_2$ from $Elements$
            $UsedClusters := UsedClusters \cup \{c_1, c_2\}$
    Repeat steps 2-4 to recalculate $Scores$

Figure 2: Resolver's **Clustering Algorithm**

## 5.1 Algorithm Analysis

Let $D$ be the set of extracted assertions. The following analysis[2] shows that one iteration of merges takes time $O(|D| \log |D|)$. Let $NC$ be the number of comparisons between strings in step 4. To simplify the analysis, we consider only those properties that contain a relation string and an argument 1 string. Let $Properties$ be the set of all such properties that apply to fewer than $Max$ strings, and let $Strings_p$ be the set of all strings that a particular

---

2. If the $Max$ parameter is allowed to vary with $\log |D|$, rather than remaining constant, the same analysis leads to a slightly looser bound that is still better than $O(|D|^2)$.

|           | Num. Strings | Compare All | Resolver | Speedup |
|-----------|--------------|-------------|----------|---------|
| Objects   | 9,797        | 47,985,706  | 352,177  | 136x    |
| Relations | 10,151       | 51,516,325  | 105,915  | 486x    |

Table 1: Resolver's clustering algorithm cuts down on the number of comparisons made between pairs of strings when clustering a data set of 2.1 million TextRunner extractions. "Compare All" lists the number of comparisons that would be made if every string were compared to every other one. Resolver reduces comparisons between object strings by a factor of 136 compared to this baseline, and comparisons between relations strings by a factor of 486.

property $p$ applies to. The number of comparisons is given by the size of the union of the set of comparisons made for each property, which is upper-bounded by the sum of the maximum number of comparisons made for each property:

$$NC = \left| \bigcup_{p \in Properties} \{pair = \{s_1, s_2\} | pair \subset Strings_p\} \right|$$

$$\leq \sum_{p \in Properties} |\{pair = \{s_1, s_2\} | pair \subset Strings_p\}|$$

$$= \sum_{p \in Properties} \frac{|Strings_p| \times (|Strings_p| - 1)}{2}$$

Since each $Strings_p$ contains at most $Max$ elements, we can upper-bound this expression by

$$NC \leq \sum_{p \in Properties} \frac{|Strings_p| \times (Max - 1)}{2}$$

$$= \frac{(Max - 1)}{2} \times \sum_{p \in Properties} |Strings_p|$$

$$\leq \frac{(Max - 1)}{2} \times |D|$$

The last step bounds $\sum_p |Strings_p|$ with $|D|$, since the number of extractions is equal to the number of times that each property is extracted. Since $\sum_p |Strings_p|$ is summing only over properties that apply to fewer than $Max$ strings, $|D|$ may be greater than this sum. Overall, the analysis shows that $NC$ is linear in $|D|$. Note that in general this bound is quite loose because most properties apply to only a small number of strings, far fewer than $Max$.

Step 5 requires time $O(|D| \log |D|)$ to sort the comparison scores and perform one iteration of merges. If the largest cluster has size $K$, in the worst case the algorithm will take

$K$ iterations (and in the best case it will take $\log K$). In our experiments, the algorithm never took more than 9 iterations.

The analysis thus far has related the computational complexity to $|D|$, the size of the input data set of extractions. Most existing techniques, however, have been analyzed in terms of $|S|$, the number of distinct strings to be clustered. In order to relate the two kinds of analysis, we observe that linguistic data naturally obeys a Zipf distribution for the frequency of its distinct strings. That is, the most common string appears many times in the extractions; the next-most common appears roughly $\left(\frac{1}{2}\right)^z$ times as often for some parameter $z$; the next most common appears roughly $\left(\frac{1}{3}\right)^z$ times as often; and so on. The parameter $z$ is known as the Zipf parameter, and for naturally-occurring text it has typically been observed to be around 1 (Zipf, 1932; Manning & Schuetze, 1999). If we can characterize the Zipf distribution for the input data set of extractions, we can rewrite the number of extractions $|D|$ in terms of the number of distinct strings $|S|$, since $|D| = \sum_{s \in S} \text{frequency}(s)$. Following this line of thought to its conclusion, we find that when $z < 1$, as it is for our data set, $|D|$ grows linearly with $|S|$, and a complexity of $O(|D| \log |D|)$ is equivalent to a complexity of $O(|S| \log |S|)$. When $z = 1$, $O(|D| \log |D|)$ is equivalent to a bound of $O(|S| \log^2 |S|)$. And when $z > 1$, the bound is $O(|S|^z \log |S|)$. Not until $z = 2$ is the asymptotic bound of $O(|S|^2 \log |S|)$ worse than the $O(|S|^2)$ bound for comparing all string pairs, and such a high value of $z$ is highly unlikely for naturally occurring text. For more details and a complete analysis, see Appendix C.

## 5.2 Relation to Other Speed-Up Techniques

McCallum, Nigam, and Ungar (2000) proposed a widely-used technique for pre-processing a data set to reduce the number of comparisons made during clustering. They use a cheap comparison metric to place objects into overlapping "canopies," and then use a more expensive metric to cluster objects appearing in the same canopy. The RESOLVER clustering algorithm is in fact an adaptation of the canopy method: like the Canopies method, it uses an index to eliminate many of the comparisons that would otherwise need to be made. Our method adds the restriction that strings are not compared when they share only high-frequency properties. The Canopy method works well on high-dimensional data with many clusters, which is the case with our problem. Our contribution has been to observe that if we restrict comparisons in a novel and well-justified way, we can obtain a new theoretical bound on the complexity of clustering text data.

The merge/purge algorithm (Hernandez & Stolfo, 1995) assumes the existence of a particular attribute such that when the data set is sorted on this attribute, matching pairs will all appear within a narrow window of one another. This algorithm is $O(M \log M)$ where $M$ is the number of distinct strings. However, there is no attribute or set of attributes that comes close to satisfying this assumption in the context of domain-independent information extraction.

RESOLVER's clustering task can in part be reduced to a task of nearest-neighbor search, for which several recent systems have developed fast new algorithms. The reduction works as follows: the nearest-neighbor retrieval techniques can be used to find the most similar string for every distinct string in the corpus, and then RESOLVER's merge criteria can decide which of these $M$ pairs to actually merge. Several of the fastest nearest-neighbor techniques perform approximate nearest-neighbor search: given an error tolerance $\epsilon$, such techniques will return a neighbor for a query node $q$ that is at most $1 + \epsilon$ times as far from $q$ as the true nearest neighbor of $q$.

Examples of nearest-neighbor techniques can be divided into those that use hash-based or tree-based indexing schemes. Locality-Sensitive Hashing uses a combination of hashing functions to retrieve approximate nearest neighbors in time $O(n^{\frac{1}{1+\epsilon}})$ for error tolerance $\epsilon$. So if for a given query point $q$ we are willing to accept neighbors that are at a distance of at most twice the distance of its true nearest neighbor ($\epsilon = 2$), then the running time will be $O(n^{\frac{1}{2}}) = O(\sqrt{n})$ to find a single nearest neighbor (Gionis, Indyk, & Motwani, 1999). More recently, tree-based index structures such as metric cover trees (Beygelzimer, Kakade, & Langford, 2006) and hybrid spill trees (Liu, Moore, Gray, & Yang, 2004), have offered competitive or even better performance than Locality-Sensitive Hashing. The tree-based algorithms have a complexity of $O(d \log n)$, where $d$ is the dimensionality of the space, to find a single nearest neighbor. Metric trees offer exact nearest-neighbor search, and spill trees offer faster search in practice at the cost of finding approximate solutions and using more space for their index. These indexing schemes are powerful tools for nearest-neighbor search, but their dependence on the dimensionality of the space makes it costly to apply them in our case. RESOLVER operates in a space of hundreds of thousands of dimensions (the number of distinct extract properties), while the fastest of these techniques have been applied to spaces of around a few thousand dimensions (Liu et al., 2004). RESOLVER determines the exact nearest neighbor, and in fact the exact distance between all relevant pairs of points under the mild assumptions stated above, while operating in a huge-dimensional space.

### 5.3 RESOLVER Implementation

RESOLVER currently exists as a Java package containing 23,338 lines of code. It has separate modules for calculating the Extracted Shared Property Model and the String Similarity Model, as well as for clustering extractions. The basic version of the system accepts a file containing tuples of strings as input, one tuple per line. Optionally, it accepts manually labeled clusters as input as well, and will use those to output precision and recall scores. The output of the system is two files containing all object clusters and relation clusters of size two or more, respectively. Optionally, the system also outputs precision and recall scores. Several other options allow the user to run extensions to the basic RESOLVER system, which are discussed below in Section 7.

RESOLVER is currently a part of the TEXTRUNNER demonstration system. The demonstration system is available for keyword searches over the Web at http://www.cs.washington.edu/research/textrunner/. This demonstration system contains extractions from several hundred million Web documents. The extractions were fed into RESOLVER and the resulting clusters were added to the TEXTRUNNER index so that keyword searches return results for any member of the cluster containing the keyword being searched for, and the displayed results are condensed such that members of the same cluster are not repeated.

## 6. Experiments

Several experiments below test Resolver and ESP, and demonstrate their improvement over related techniques in paraphrase discovery, $s_{MI}$ (Lin & Pantel, 2001) and the Cosine Similarity Metric (CSM) (Salton & McGill, 1983; Hasegawa et al., 2004; Mann & Yarowsky, 2003). The first experiment compares the performance of the various similarity metrics, and shows that Resolver's output clusters are significantly better than ESP's or SSM's, and that ESP's clusters are in turn significantly better than $s_{MI}$'s or CSM's. The second experiment measures the sensitivity of the ESP model to its hidden parameter, and shows that for a very wide range of parameter settings, it is able to outperform both the $s_{MI}$ and CSM models.

### 6.1 Experimental Setup

The models are tested on a data set of 2.1 million assertions extracted from a Web crawl. All models run over all assertions, but compare only those objects or relations that appear at least 25 times in the data, to give the distributional similarity models sufficient data for estimating similarity. Although this restriction limits the applicability of Resolver, we note that it is intuitive that this should be necessary for unsupervised clustering, since such systems by definition start with no knowledge about a string. They must see some number of examples before it is reasonable to expect them to make decisions about them. We also note that Downey, Schoenmackers, and Etzioni (2007) have shown for a different problem how bootstrapping techniques can leverage performance on high-frequency examples to build accurate models for low-frequency items.

Only proper nouns[3] are compared, and only those relation strings that contain no punctuation or capital letters are compared. This helps to restrict the experiment to strings that are less prone to extraction errors. However, the models do use the other strings as features. In all, the data contains 9,797 distinct proper object strings and 10,151 distinct proper relation strings that appear at least 25 times. We created a gold standard data set by manually clustering a subset of 6,000 object and 2,000 relation strings. In total, our gold standard data sets contains 318 true object clusters and 330 true relation clusters with at least 2 elements each.

As noted previously (Section 3.1), polysemous strings pose a particular representational trouble for creating a gold standard data set, since there is no correct clustering that captures all of the synonymy relationships for polysemous strings, in general. We adopted the following data-oriented strategy: polysemous strings were not clustered with other strings unless there was a match for every sense of the strings that appeared in the data. For example, there have been two U.S. Presidents named "Roosevelt": Theodore Roosevelt and Franklin Delano Roosevelt. After applying the criterion above, the gold standard data contained a cluster for `FDR` and `President Franklin Roosevelt`, since both referred to Franklin Delano Roosevelt unambiguously in this dataset. Likewise, `President Theodore Roosevelt` and `Teddy Roosevelt` were put into their own cluster. The terms `Roosevelt` and `President Roosevelt`, however, were used in various places to refer to both men, and so they could not be clustered with either the Franklin Roosevelt cluster or the Theodore Roosevelt cluster. Since they had the same set of senses in the data, the gold standard contained a separate cluster containing just these two strings. Section 8.2 describes an extension to RESOLVER that handles polysemous names. Our criterion for polysemy prevented 480 potential merges in our gold standard data set between object clusters that might be synonymous. The prevented merges usually affected acronyms, first names, and words like "Agency" that might refer to a number of institutions, and they represent less than 10% of the strings in the gold standard object data set.

---

3. The following heuristic was used to detect proper nouns: if the string consisted of only alphabetic characters, whitespace, and periods, and if the first character of every word is capitalized, it is considered a proper noun. Otherwise, it is not.

In addition to a gold standard data set for evaluation, we manually created a data set of development data containing 5 correct pairs of objects, 5 correct pairs of relations, and also 5 examples of incorrect pairs for each. These 20 examples were not used in the evaluation data. The development data was used to estimate a value for the ESP model's hidden parameter $N$, called its property multiple (see Section 4.2). We used a simple hill-climbing search procedure to find a value for $N$ separately for objects and relations, and found that $N = 30$ worked best for objects on development data, and $N = 500$ for relations. Although the amount of data required to set this parameter effectively is very small, it is nevertheless an important topic for future work to come up with a method that will estimate this parameter in a completely unsupervised manner in order to fully automate RESOLVER.

For our comparisons, we calculated the Cosine Similarity Metric (CSM) using the technique described in Section 4.3 and Equation 5, and the $s_{MI}$ metric as defined in Equation 4.

## 6.2 Clustering Analysis

Our first experiment compares the precision and recall of clusterings output by five similarity metrics: two kinds of previous work in paraphrase discovery, CSM and $s_{MI}$; two components of RESOLVER, ESP and SSM; and the full RESOLVER system.

The precision and recall of a clustering is measured as follows: hypothesis clusters are matched with gold clusters such that each hypothesis cluster matches no more than one gold cluster, and *vice versa*. This mapping is computed so that the number of elements in hypothesis clusters that intersect with elements in the matching gold clusters is maximized. All such intersecting elements are marked correct. Any elements in a hypothesis cluster that do not intersect with the corresponding gold cluster are marked incorrect, or irrelevant if they do not appear in the gold clustering at all. Likewise, gold cluster elements are marked as *found* if the matching hypothesis cluster contains the same element, or *not found* otherwise. The precision is defined as the number of correct hypothesis elements in clusters containing at least two relevant (correct or incorrect) elements, divided by the total number of relevant hypothesis elements in clusters containing at least two relevant items. The recall is defined as the number of found gold elements in gold clusters of size at least two, divided by the total number of gold elements in clusters of size at least two. We consider only clusters of size two or more in order to focus on the interesting cases.

Each model requires a threshold parameter to determine which scores are suitable for merging. For these experiments we arbitrarily chose a threshold of 3 for the ESP model (that is, the data needs to be 3 times more likely given the merged cluster than the unmerged clusters in order to perform the merge) and chose thresholds for the other models by hand so that the difference between them and ESP would be roughly even between precision and recall, although for relations it was harder to improve the recall. Table 2 shows the precision and recall of our models.

|        | Objects | | | Relations | | |
| --- | --- | --- | --- | --- | --- | --- |
| Model | Prec. | Rec. | F1 | Prec. | Rec. | F1 |
| CSM | 0.51 | 0.36 | 0.42 | 0.62 | 0.29 | 0.40 |
| $s_{MI}$ | 0.52 | 0.38 | 0.44 | 0.61 | 0.28 | 0.38 |
| ESP | **0.56** | 0.41 | 0.47 | **0.79** | 0.33 | 0.47 |
| SSM | **0.62** | 0.53 | 0.57 | **0.85** | 0.25 | 0.39 |
| Resolver | **0.71** | 0.66 | 0.68 | **0.90** | **0.35** | 0.50 |

Table 2: **Comparison of the cosine similarity metric (CSM), $s_{MI}$, Resolver components (SSM and ESP), and the Resolver system.** Bold indicates the score is significantly different from the score in the row above at $p < 0.05$ using the chi-squared test with one degree of freedom. Using the same test, Resolver is also significantly different from ESP, $s_{MI}$, and CSM in recall on objects, and from $s_{MI}$, CSM and SSM in recall on relations. Resolver's F1 on objects is a 19% increase over SSM's F1. Resolver's F1 on relations is a 28% increase over SSM's F1. No significance tests were performed on the F1 values.

### 6.3 Sensitivity Analysis

The ESP model requires a parameter for the number of potential properties of a string, but the performance of ESP is not strongly sensitive to the exact value of this parameter. As described in Section 4.2, we assume that the number of potential properties is a multiple $N$ of the number of extractions for a string. In the above experiments, we chose values of $N = 30$ for objects and $N = 500$ for relations, since they worked well on held-out data. However, as Tables 3 and 4 show, the actual values of these parameters may vary in a large range, while still enabling ESP to outperform $s_{MI}$ and CSM.

In these experiments, we measured precision and recall for just the similarity metrics, without performing any clustering. We used the similarity metrics to sort the pairs of strings (but only those pairs that share at least some property) in descending order of similarity. We then place a threshold $T$ on the similarity, and measure precision as the number of correct synonym pairs with similarity greater than $T$ divided by the total number of pairs with similarity greater than $T$. We measure recall by the number of correct synonym pairs with similarity greater than $T$ divided by the total number of correct synonym pairs. By varying $T$, we can create a precision-recall curve and measure the area underneath the curve.

These tables highlight two significant results. First, for both objects and relations the ESP model outperforms CSM and $s_{MI}$ by a large amount for parameter settings that vary by close to a factor of two in either direction from the value we determined on development data. Thus although we required a small amount of data to determine a value for this parameter, the performance of ESP is not overly sensitive to the exact value. Second, the ESP model clearly provides a significant boost to the performance of the SSM model, as RESOLVER's performance significantly improves over SSM's.

| Metric | AUC | Fraction of Max. AUC | Improvement over Baseline |
|--------|-----|----------------------|---------------------------|
| CSM | 0.0061 | 0.011 | -21% |
| $s_{MI}$ | 0.0083 | 0.014 | 0% |
| ESP-10 | 0.019 | 0.033 | 136% |
| ESP-30 | 0.024 | 0.041 | **193%** |
| ESP-50 | 0.022 | 0.037 | 164% |
| ESP-90 | 0.018 | 0.031 | 121% |
| SSM | 0.18 | 0.31 | 0% |
| Resolver | 0.22 | 0.38 | **23%** |

Table 3: **Area Under the precision-recall Curve (AUC) for object synonymy. The ESP model significantly outperforms $s_{MI}$ and CSM in AUC for a wide range of parameter settings. Likewise, Resolver significantly outperforms SSM in AUC.** The maximum possible AUC is less than one because many correct string pairs share no properties, and are therefore not compared by the clustering algorithm. The third column shows the score as a fraction of the maximum possible area under the curve, which for objects is 0.57. The improvement over baseline column shows how much the ESP curves improve over $s_{MI}$, and how much Resolver improves over SSM.

| Metric | AUC | Fraction of Max. AUC | Improvement over Baseline |
|--------|-----|----------------------|---------------------------|
| CSM | 0.0035 | 0.034 | -19% |
| $s_{MI}$ | 0.0044 | 0.042 | 0% |
| ESP-50 | 0.0048 | 0.046 | 9.5% |
| ESP-250 | 0.0087 | 0.083 | 98% |
| ESP-500 | 0.0096 | 0.093 | 121% |
| ESP-900 | 0.010 | 0.098 | **133%** |
| SSM | 0.022 | 0.24 | 0% |
| Resolver | 0.029 | 0.31 | **31%** |

Table 4: **Area Under the precision-recall Curve (AUC) for relation synonymy. The ESP model significantly outperforms $s_{MI}$ and CSM in AUC for a wide range of parameter settings. Likewise, Resolver significantly outperforms SSM in AUC.** The maximum possible area is less than one because many correct string pairs share no properties, and are therefore not compared by the clustering algorithm. The third column shows the score as a fraction of the maximum possible area under the curve, which for relations is 0.094. The improvement over baseline shows how much the ESP curves improve over $s_{MI}$, and how much Resolver improves over SSM.

## 6.4 Discussion

In all experiments, ESP outperforms both CSM and $s_{MI}$. The sensitivity analysis shows that this remains true for a wide range of hidden parameters for ESP, for both objects and relations. Moreover, ESP's improvement over the comparison metrics holds true when the metrics are used in clustering the data. $s_{MI}$'s performance is largely the same as CSM in every experiment. Somewhat surprisingly, $s_{MI}$ performs worse on relation clustering than on object clustering, even though it is designed for relation similarity.

The results show that the three distributional similarity models perform below the SSM model on its own for both objects and relations, both in the similarity experiments and the clustering experiments. The one exception is in the clustering experiment for relations, where SSM had a poor recall, and thus had lower F1 score than ESP and CSM. This is to be expected, since ESP, $s_{MI}$, and CSM make predictions based on a very noisy signal. For example, `Canada` shares more properties with `United States` in our data than `U.S.` does, even though `Canada` appears less often than `U.S.`. Importantly, though, there is a significant improvement in both precision and recall when using a combined model over using SSM alone. RESOLVER's F1 is 19% higher than SSM's on objects, and 28% higher on relations in the clustering experiments.

Interestingly, the distributional similarity metrics (ESP, $s_{MI}$, and CSM) perform significantly worse in the task of ranking string pairs than in the clustering task. One reason is that the task of ranking string pairs does not measure performance when comparing a cluster of two strings against a cluster of two other strings. In a greedy clustering process such as the one used by RESOLVER, large groups of correct clusters can be formed as long as the similarity metrics rank some correct pair of strings near the top, and are able to improve their estimates of similarity when comparing clusters. This issue requires further investigation.

There is clearly room for improvement on the synonym resolution task. Error analysis shows that most of RESOLVER's mistakes are due to three kinds of errors:

1. *Extraction errors.* For example, `US News` gets extracted separately from `World Report`, and then RESOLVER clusters them together because they share almost all of the same properties.

2. *Similarity vs. Identity.* For example, `Larry Page` and `Sergey Brin` get merged, as do `Angelina Jolie` and `Brad Pitt`, and `Asia` and `Africa`.

3. *Multiple word senses.* For example, there are two President Bushes; also, there are many terms like `President` and `Army` that can refer to multiple distinct entities.

Extraction systems are improving their accuracy over time, and we do not further address these errors. The next two sections develop techniques to address the second and third of these kinds of errors, respectively.

## 7. Similar and Identical Pairs

As the error analysis above suggests, similar objects that are not exact synonyms make up a large fraction of RESOLVER's errors. This section describes three techniques for dealing with such errors.

For example, RESOLVER is likely to make a mistake with the pair `Virginia` and `West Virginia`. They share many properties because they have the same type (U.S. states), and they have high string similarity. Perhaps the easiest approach for determining that these two are not synonymous is simply to collect more data about them. While they are highly similar, they will certainly not share all of their properties; they have different governors, for example. However, for highly similar pairs such as these two, the amount of data required to decide that they are not identical may be huge, and simply unavailable.

Fortunately, there are more sophisticated techniques for making decisions with the available data. One approach is to consider the distribution of words that occur between candidate synonyms. Similar words are likely to be separated by conjunctions (e.g., "Virginia and West Virginia") and domain-specific relations that hold between two objects of the same type (e.g., "Virginia is larger than West Virginia"). On the other hand, synonyms are more likely to be separated by highly specialized phrases such as "a.k.a." Section 7.1 describes a method for using this information to distinguish between similar and identical pairs.

A second approach is to consider how candidate synonyms behave in the context of relations with special distributions, like functions or inverse functions. For example, the "$x$ is capital of $y$" relation is an inverse function: every $y$ argument has at most one $x$ argument[4]. If capitals are extracted for both West Virginia and Virginia, then they may be ruled out as a synonymous pair when the capitals are seen to be different. On the other hand, if `Virginia` and `VA` share the same capital, that is much stronger evidence that the two are the same than if they shared some other random property, such as that a town called `Springfield` is located there. Section 7.2 describes a method for eliminating similar pairs because they have different values for the same function or inverse function, and Section 7.3 illustrates a technique for assigning different weights to different evidence based on how close to functional the property is. Section 7.4 gives results for each of these techniques.

### 7.1 Web Hitcounts for Synonym Discovery

While names for two similar objects may often appear together in the same sentence, it is relatively rare for two different names of the same object to appear in the same sentence. Moreover, synonymous pairs tend to appear in idiosyncratic contexts that are quite different from the contexts seen between similar pairs. RESOLVER exploits this fact by querying the Web to determine how often a pair of strings appears together in certain contexts in a large corpus. When the hitcount is high, RESOLVER can prevent the merge.

Specifically, given a candidate synonym pair $s_1$ and $s_2$, the Coordination-Phrase Filter uses a discriminator phrase (Etzioni et al., 2005) of the form "$s_1$ and $s_2$". It then computes

---

[4]. It is also a function.

a variant of pointwise mutual information, given by

$$\text{coordination score}(s_1, s_2) = \frac{\text{hits}(s_1 \text{ and } s_2)^2}{\text{hits}(s_1) \times \text{hits}(s_2)}$$

The filter removes from consideration any candidate pair for which the coordination score is above a threshold, which is determined on a small development set. The results of coordination-phrase filtering are presented below.

The Coordination-Phrase Filter uses just one possible context between candidate synonym pairs. A simple extension is to use multiple discriminator phrases that include common context phrases like "or" and "unlike." A more complex approach could measure the distribution of words found between a candidate pair, and compare that distribution with the distributions found between known similar or known identical pairs. These are important avenues for further investigation.

One drawback of this approach is that it requires text containing a pair of objects in close proximity. For a pair of rare strings, such data will be extremely unlikely to occur — this type of test exacerbates the data sparsity problem. The following two sections describe two techniques that do not suffer from this particular problem.

## 7.2 Function Filtering

Functions and inverse functions can help to distinguish between similar and identical pairs. For example, `Virginia` and `West Virginia` have different capitals: respectively, `Richmond` and `Charleston`. If both of these facts are extracted, and if RESOLVER knows that the `capital of` relation is an inverse function, it ought to prevent `Virginia` and `West Virginia` from merging.

Given a candidate synonym pair $x_1$ and $x_2$, the Function Filter prevents merges between strings that have different values for the same function. More precisely, it decides that two strings $y_1$ and $y_2$ *match* if their string similarity is above a high threshold. It prevents a merge between $x_1$ and $x_2$ if there exists a function $f$ and extractions $f(x_1, y_1)$ and $f(x_2, y_2)$, and there are no such extractions such that $y_1$ and $y_2$ match (and *vice versa* for inverse functions). Experiments described in Section 7.4 show that the Function Filter can improve the precision of RESOLVER without significantly affecting its recall.

The Function Filter requires knowledge about which relations are actually functions or inverse functions. Others have investigated techniques for determining such properties of relations automatically (Popescu, 2007); in the experiments, a pre-defined list of functions is used. Table 5 lists the set of functions used in the experiments for the Function Filter. These functions were selected by manually inspecting a set of 500 common relations from TEXTRUNNER's extractions, and selecting those that were reliably functional. Only a few met the criteria, partly because of polysemy in the data, and partly because of extraction noise.

## 7.3 Function Weighting

While the Function Filter uses functions and inverse functions as negative evidence, it is also possible to use them as positive evidence. For example, the relation `married` is not strictly one-to-one, but for most people the set of spouses is very small. If a pair of strings are extracted with the same spouse—*e.g.*, FDR and `President Roosevelt` share the property (`married, Eleanor Roosevelt`)—this is far stronger evidence that the two strings are identical than if they shared some random property, such as (`spoke to, reporters`).

| is capital of | is capital city of |
| named after | was named after |
| headquartered in | is headquartered in |
| was born in | was born on |

Table 5: **The set of functions used by the Function Filter.**

There are several possibilities for incorporating this insight into RESOLVER. First, any such technique will need some method for estimating the "function-ness" of a property, or how close the property is to being functional. We define the *degree* of a relation to be the number of $y$ values that are expected to hold true for a given $x$ value. We call a property *high-degree* if it is expected to apply to many strings (highly non-functional), and *low-degree* if it is expected to apply to few strings (close to functional).

The degree of a property may be estimated from the relation involved in the property and the set of extractions for that relation, or it may be based on how many objects the property applies to. For example, if there are 100 unique extractions for the `married` relation, and there are 80 unique $x$ argument strings in those 100 extractions, then on average each $x$ string participates in $100/80 = 1.25$ `married` relations. One method might assign every property containing the `married` relation this statistic as the degree. On the other hand, suppose there are two extractions for the property (`married, John Smith`). A second method is to assign a degree of 2 to this property.

There are also two possible ways to incorporate the degree information into the ESP model. The ESP model may be altered so that it directly models the degrees of the properties during the process of selecting balls from urns, but this vastly complicates the model and may make it much more computationally expensive. A second option is to reweight the number of shared properties between strings based on a TF-IDF style weighting of the properties, and calculate the ESP model using this parameter instead. This requires modifying the ESP model so that it can handle non-integer values for the number of shared properties.

In experiments so far, one set of these options was explored, while others remain for future investigation. The Weighted Extracted Shared Property Model (W-ESP) sets the degree of a property to be the number of extractions for that property. Second, if strings $s_i$ and $s_j$ share all properties $p \in P$, it sets the value for the number of shared properties between $s_i$ and $s_j$ to be

$$\sum_{p \in P} \frac{1}{\text{degree}(p)}$$

The ESP model has been changed to handle continuous values for the number of shared properties by changing all factorials to gamma functions, and using Stirling's approximation whenever possible.

| Model | Prec. | Rec. | F1 |
|---|---|---|---|
| Resolver | 0.71 | 0.66 | 0.68 |
| Resolver + Function Filtering | 0.74 | 0.66 | 0.70 |
| Resolver + Coordination Phrase Filtering | **0.78** | 0.68 | 0.73 |
| Resolver + Weighted ESP | 0.71 | 0.65 | 0.68 |
| Resolver + Function and Coord. Phrase Filtering | **0.78** | 0.68 | 0.73 |

Table 6: **Comparison of object merging results for the Resolver system, Resolver plus Function Filtering, Resolver plus Coordination-Phrase Filtering, Resolver using the Weighted Extracted Shared Property Model, and Resolver plus both types of filtering.** Bold indicates the score is significantly different from Resolver's score at $p < 0.05$ using the chi-squared test with one degree of freedom. Resolver+ Coordination Phrase Filtering's F1 on objects is a 28% increase over SSM's F1, and a 7% increase over Resolver's F1.

Unlike the Function Filter, the W-ESP model does not require additional knowledge about which relations are functional. And unlike the Coordination-Phrase Filter, it does not require Web hitcounts or a training phase. It works on extracted data, as is.

### 7.4 Experiments

The extensions to Resolver attempt to address the confusion between similar and identical pairs. Experiments with the extensions, using the same datasets and metrics as in Section 6 demonstrate that the Function Filter (FF) and the Coordination-Phrase Filter (CPF) boost Resolver's precision. Unfortunately, the W-ESP model yielded essentially no improvement of Resolver.

Table 6 contains the results of our experiments. With coordination-phrase filtering, Resolver's F1 is 28% higher than SSM's on objects, and 6% higher than Resolver's F1 without filtering. While function filtering is a promising idea, FF provides a smaller benefit than CPF on this dataset, and the merges that it prevents are, with a few exceptions, a subset of the merges prevented by CPF. This is in part due to the limited number of functions available in the data.

Both the Function Filter and the Coordination-Phrase Filter consistently blocked merges between highly similar countries, continents, planets, and people in our data, as well as some other smaller classes. The biggest difference is that CPF more consistently has hitcounts for the similar pairs that tend to be confused with identical pairs. Perhaps as the amount of extracted data grows, more functions and extractions with functions will be extracted, allowing the Function Filter to improve.

Part of the appeal of the W-ESP model is that it requires none of the additional inputs that the other two models require, and it applies to each property, rather than to a subset of the relations like the Function Filter. Like TFIDF weighting for the Cosine Similarity Metric, the W-ESP model uses information about the distribution of the properties in the data to weight each property. For the data extracted by TextRunner, neither W-ESP nor TFIDF weighting seems to have a positive effect. More experiments are required to test whether W-ESP might prove more beneficial on other data sets where TFIDF does have a positive effect.

## 8. RESOLVER and Cross-Document Entity Resolution

Up to this point, we have made the single-sense assumption, or the assumption that every token has exactly one meaning. While this assumption is defensible in small domains, where named entities and relations rarely have multiple meanings, even there it can cause problems: for example, the names Clinton and Bush each refer to two major players in American politics, as well as a host of other people. When extractions are taken from multiple domains, this assumption becomes more and more problematic.

We now describe a refinement of the RESOLVER system that handles the task of Cross-Document Entity Resolution (Bagga & Baldwin, 1998), in which tokens or names may have multiple referents, depending on context. An experiment below compares RESOLVER with an existing entity resolution system (Li et al., 2004a), and demonstrates that RESOLVER can handle polysemous named entities with high accuracy. This extension could theoretically be applied to highly polysemous tokens such as common nouns, but this has not yet been empirically demonstrated.

### 8.1 Clustering Polysemous Names with RESOLVER

Recall that the synonym resolution task is defined as finding clusters in the set of distinct strings $S$ found in a set of extractions $D$ (Section 3). Cross-Document Entity Resolution differs from synonym resolution in that it requires a clustering of the set of all string *occurrences*, rather than the set of distinct strings. For example, suppose a document contains two occurrences of the token DP, one where it means "Design Pattern" and one where it means "Dynamic Programming." Synonym resolution systems treat DP as a single item, and will implicitly cluster both occurrences of DP together. A Cross-Document Entity Resolution system treats each occurrence of DP separately, and therefore has the potential to put each occurrence in a separate cluster when they mean different things. In this way, a Cross-Document Entity Resolution system has the potential to handle polysemous names correctly.

Because of the change in task definition, the sources of evidence for similarity are sparser. For each occurrence of a named entity in its input, RESOLVER has just a single TEXT-RUNNER extraction describing the occurrence. To achieve reasonable performance, it needs more information about the context in which a named entity appears. We change RESOLVER's representation of entity occurrences to include the nearest $E$ named entities in the text surrounding the occurrence. That is, each entity occurrence $x$ is represented as a set of named entities $y$, where $y$ appears among the nearest $E$ entities in the text surrounding $x$. Suppose, for instance, that $e_1$ is an occurrence of DP with Bellman and Viterbi in its context, and $e_2$ is another occurrence with OOPSLA and Factory in its context. $e_1$ would be represented by the set {Bellman, Viterbi}, and $e_2$ would be represented by the set {Factory, OOPSLA}.

Table 7 summarizes the major ways in which we extended RESOLVER to handle polysemous names. With these extensions in place, RESOLVER can proceed to cluster occurrences of entities more or less the same way that it clusters entity names for synonym resolution.

|  | Original RESOLVER | Extended RESOLVER |
|---|---|---|
| Input | A set of distinct strings $S$, and for each $s \in S$, a set of extracted properties of $s$. | A bag of string occurrences $O$, and for each occurrence $o \in O$, a set of named entities appearing close to $o$ in context. |
| SSM Compares | Character sequences | Character sequences |
| ESP Compares | Sets of extracted properties | Sets of named entities |
| Output | A clustering of the set of distinct strings $S$ | A clustering of the set of string occurrences $O$ |

Table 7: **The differences between the original** RESOLVER **system and the extended** RESOLVER **system for handling polysemous names.**

The SSM model works as above, and the ESP model calculates probabilities of coreference based on sets of named entities in context rather than extracted properties. The clustering algorithm eliminates comparisons between occurrences that share no part of their contexts, or only very common contextual elements. In the end, RESOLVER produces sets of coreferential entity occurrences, which can be used to annotate extractions containing these entity occurrences for coreference relationships.

## 8.2 Experiment with Cross-Document Entity Resolution

We tested RESOLVER's ability to handle polysemous names on a data set of 300 documents from 1998-2000 New York Times articles in the TREC corpus (Voorhees, 2002). Li et al. (2004b) automatically ran a named-entity tagger on these documents and manually corrected them to identify approximately 4,000 occurrences of people's names. They then manually annotated the occurrences to form a gold standard set of coreferential clusters.

For each named entity occurrence in this data set, we extracted the set of the closest $E$ named entities, with $E$ set to 100, to represent the context for the named entity occurrence. We then ran RESOLVER to cluster the entity occurrences. We set ESP's latent parameter $N$ to 30, as in the experiments above. We did not have any development data to set the merge threshold, so we used the following strategy: we arbitrarily picked a single occurrence of a common name from this data set (`Al Gore`), found a somewhat uncommon variant of the name (`Vice President Al Gore`), and set the threshold at a value just below the similarity score for this pair (7.5). For every round of merging in RESOLVER's clustering algorithm, we filtered the top 20 proposed merges using the Coordination Phrase Filter, with the same threshold as used in the previous experiments.

Li *et al.* propose a generative model of entity coreference that we compare against. Their model requires databases of information about titles, first names, last names, genders, nicknames, and common transformations of these attributes of people's names to help compute the probability of coreference. It uses Expectation-Maximization over the given data set to compute parameters, and an inference algorithm that is $O(N^2)$ in the number of word occurrences $N$. Full details are provided by Li et al. (2004b).

|  | Precision | Recall | F1 |
|---|---|---|---|
| RESOLVER | 97.3 | 94.7 | 96.0 |
| Li *et al.* | 91.5 | 94.0 | 92.7 |

Table 8: RESOLVER **outperforms the system by Li** *et al.* **on a Cross-Document Entity Resolution task involving polysemous people's names. Differences are statistically significant for both precision and recall using the two-tailed Chi-Square test with one degree of freedom** ($p < 0.01, \alpha = 910.9$ **for precision and** $\alpha = 20.6$ **for recall).**

Following Li *et al.*, we evaluate clusters using precision and recall calculated as follows: let $O_p$ be the set of entity occurrence pairs that are predicted to be coreferential (*i.e.*, they belong to the same cluster), and let $O_a$ denote the set of correct coreferential pairs, as calculated from the manual clustering. Then precision $P = \frac{|O_p \cap O_a|}{|O_p|}$, recall $R = \frac{|O_p \cap O_a|}{|O_a|}$, and $F_1 = \frac{2PR}{P+R}$.

Table 8 shows the results of running RESOLVER on this data set, as well as the best results reported by Li et al. (2004b) on the same data. [5] In follow-up work, Li et al. (2004a) demonstrate that their unsupervised model outperforms three supervised techniques that learn parameters for how much different attributes (first name, honorifics, *etc.*) contribute to the similarity of occurrence pairs.

In terms of absolute performance, RESOLVER is quite accurate in dealing with the polysemous names in this data set. Its performance on this data set is significantly higher than on TEXTRUNNER extractions, partly because it has extra information available in terms of the contexts of occurrences, and partly because it is starting out with manually labelled named entities, rather than noisy extractions.

RESOLVER's precision is significantly higher than Li *et al.*'s, with roughly equal recall. Because of the large sample sizes, the differences in precision and recall are both statistically significant (two-tailed Chi-Square test with one degree of freedom, $p < 0.01$). In comparison with Li *et al.*'s system, RESOLVER's SSM model is much less sophisticated, but it compensates by using Web data and a strong measure of distributional similarity. It does not need to rely on manually curated databases for expert knowledge about the domain, or in this case, the similarity of people's names.

## 9. Conclusion and Future Work

We have shown that the unsupervised and scalable RESOLVER system is able to find clusters of coreferential object names in extracted relations with a precision of 78% and a recall of 68% with the aid of coordination-phrase filtering, and can find clusters of coreferential relation names with precision of 90% and recall of 35%. We have demonstrated significant improvements over using existing similarity metrics for this task by employing a novel probabilistic model of synonymy. On a much cleaner set of extractions from the TREC corpus, we demonstrated that RESOLVER was able to achieve 97% precision and 95% recall by employing an extension that allowed it to cluster different senses of the same name into different groups.

---

5. In follow-up work, Li et al. (2004a) report an $F_1$ score of 95.1 for this task using what appears to be the same model and the same data, but the result is calculated by testing the model on 6 random splits of the data and averaging the score. We do not have access to these random splits. One possible reason that the reported results are different is that splitting up the test data reduces the number of coreference relations that need to be found and the potential number of incorrect coreference relations that can cause a system confusion.

Perhaps the most critical aspect to extending RESOLVER is refining its ability to handle polysemy. Further experiments are needed to test its ability to handle new types of polysemous named entities and extracted data that has not been manually cleaned, as is the case for Li *et al.*'s data. In addition, we plan to incorporate the ESP model into a system for unsupervised coreference resolution, including common nouns and pronouns. We will extend the model to include aspects of local salience, which can be important in coreference decisions for noun phrases other than proper names.

Currently we are setting the ESP model's single hidden parameter using a development set. While the required amount of data is very small, the model might be more accurate and easier to use if the hidden parameter were set by sampling the data, rather than using a development set that must be manually assembled. That is, RESOLVER could inspect a substantial portion of the data, and then measure how often new properties appear in the remaining data. The rate of appearance of new properties should offer a strong signal for how to set the hidden parameter.

Several extensions to RESOLVER have dealt with ruling out highly similar non-synonyms (Section 7), with varying degrees of success at boosting RESOLVER's precision. We have also considered another extension to RESOLVER that seeks to use "mutual recursion" to boost recall, much like semi-supervised information extraction techniques use "mutual bootstrapping" between entities and patterns to increase recall (Riloff & Jones, 1999). The method begins by clustering objects, then clusters relations using the merged object clusters as properties (rather than the raw object strings), then clusters objects again using relation clusters as properties, and so on. Although we have so far been unable to boost the performance of RESOLVER using this technique on TEXTRUNNER data, experiments on artificial simulations suggest that under suitable conditions, mutual recursion could boost recall by as much as 16%. It remains an important area of future work to determine if there is natural data for which this technique is indeed useful, and to investigate other methods for increasing RESOLVER's recall.

## Acknowledgments

This research was supported in part by Temple University, NSF grants IIS-0535284 and IIS-0312988, ONR grant N00014-08-1-0431 as well as gifts from Google, and was carried out at the University of Washington's Turing Center and Temple University's Center for Information Science and Technology. We would like to thank the anonymous reviewers and the JAIR associate editor in charge of this paper for their helpful comments and suggestions. We would also like to thank the KnowItAll group at the University of Washington for their feedback and support.

## Appendix A. Derivation of the Extracted Shared Property Model

The Extracted Shared Property (ESP) Model is introduced in Section 4. It is a method for calculating the probability that two strings are synonymous, given that they share a certain number of extractions in a data set. This appendix gives a derivation of the model.

Let $s_i$ and $s_j$ be two strings, each with a set of extracted properties $E_i$ and $E_j$. Let $U_i$ and $U_j$ be the set of potential properties for each string, contained in their respective urns. Let $S_{i,j}$ be the number of properties shared between the two urns, or $|U_i \cap U_j|$. Let $R_{i,j}$ be the random variable for the synonymy relationship between $s_i$ and $s_j$, with $R_{i,j} = R_{i,j}^t$ denoting the event that they are, and $R_{i,j}^f$ that they are not. The ESP model states that the probability of $R_{i,j}^t$ is the probability of selecting the observed number of matching properties from two urns containing all matching properties, divided by the probability of selecting the observed number of matching properties from two urns which may contain some matching and some non-matching properties:

**Proposition 2** *If two strings $s_i$ and $s_j$ have $|U_i| = P_i$ and $|U_j| = P_j$ potential properties (or instances), with $min(P_i, P_j) = P_{min}$; and they appear in extracted assertions $E_i$ and $E_j$ such that $|E_i| = n_i$ and $|E_j| = n_j$; and they share $k$ extracted properties (or instances), the probability that $s_i$ and $s_j$ co-refer is:*

$$P(R_{i,j}^t | E_i, E_j, P_i, P_j) =$$

$$\frac{\binom{P_{min}}{k} \sum_{r,s \geq 0} \binom{S_{i,j}-k}{r+s}\binom{r+s}{r}\binom{P_i-P_{min}}{n_i-(k+r)}\binom{P_j-P_{min}}{n_j-(k+s)}}{\sum_{k \leq S_{i,j} \leq P_{min}} \binom{S_{i,j}}{k} \sum_{r,s \geq 0} \binom{S_{i,j}-k}{r+s}\binom{r+s}{r}\binom{P_i-S_{i,j}}{n_i-(k+r)}\binom{P_j-S_{i,j}}{n_j-(k+s)}} \tag{10}$$

The ESP model makes several simplifying assumptions:

1. Balls are drawn from the urns without replacement.

2. Draws from one urn are independent of draws from any other urn.

3. Each ball for a string is equally likely to be selected from its urn: if $U = \{u_1, \ldots, u_m\}$ and $X$ denotes a random draw from $U$, $P(X = u_i) = \frac{1}{|U|}$ for every $u_i$.

4. The prior probability for $S_{i,j}$, given the number of properties in $U_i$ and $U_j$, is uniform: $\forall_{0 \leq s \leq \min(P_i, P_j)} P(S_{i,j} = s | P_i, P_j) = \frac{1}{\min(P_i, P_j)+1}$

5. Given extracted properties for two strings and the number of potential properties for each, the probability of synonymy depends only on the number of extracted properties for each, and the number of shared properties in the extractions:
   $P(R_{i,j}^t | E_i, E_j, P_i, P_j) = P(R_{i,j}^t | k, n_i, n_j, P_i, P_j)$.

6. Two strings are synonymous if and only if they share as many potential properties as possible: $R_{i,j}^t \equiv (|U_i \cap U_j| = \min(P_i, P_j))$.

Before proving Proposition 2, we prove a simple property of urns under the assumptions above.

**Lemma 1** *Given $n$ draws without replacement from an urn containing a set of properties $U$, the probability of selecting a particular set $S \subset U$ is $\frac{1}{\binom{|U|}{|S|}}$ if $|S| = n$, and zero otherwise.*

*Proof of Lemma 1:* Let $U = \{u_1, \ldots, u_m\}$ denote the elements of $U$, and let $X_1, \ldots, X_n$ denote the independent draws from the urn. If $n = 1$, then $P(S = \{u_i\}) = P(X_1 = u_i) = \frac{1}{|U|}$ by assumption 3 above. Now suppose that $n = n_0$, and that the lemma holds for every $n' < n_0$.

$$
\begin{aligned}
P(S = \{x_1, \ldots, x_{n_0} | x_i \in U\}) &= \sum_i P(S^{n_0-1} = \{x_1, \ldots, x_{i-1}, x_{i+1}, \ldots, x_{n_0}\}) \times \\
&\quad P(X_n = x_i) \\
&= \sum_i \frac{1}{\binom{|U|}{n_0-1}} \frac{1}{|U| - n_0 + 1} \\
&= \sum_i \frac{(n_0-1)!(|U| - n_0 + 1)!}{|U|!} \frac{1}{|U| - n_0 + 1} \\
&= \frac{n_0(n_0-1)!(|U| - n_0 + 1)(|U| - n_0)!}{|U|!(|U| - n_0 + 1)} \\
&= \frac{n_0!(|U| - n_0)!}{|U|!} \\
&= \frac{1}{\binom{|U|}{n_0}}
\end{aligned}
$$

$\square$

*Proof of Proposition 2:*

We begin by transforming the desired expression, $P(R_{i,j}^t | E_i, E_j, P_i, P_j)$, into something that can be derived from the urn model. By assumptions 5 and 6, we get

$$P(R_{i,j}^t | E_i, E_j, P_i, P_j) = P(S_{i,j} = P_{min} | k, n_i, n_j, P_i, P_j) \tag{11}$$

Then, by applying Bayes Rule, we get

$$P(S_{i,j} = P_{min} | k, n_i, n_j, P_i, P_j) =$$

$$\frac{P(k | S_{i,j} = P_{min}, n_i, n_j, P_i, P_j) P(S_{i,j} = P_{min} | n_i, n_j, P_i, P_j)}{\sum_{k \le S_{i,j} \le P_{min}} P(k | n_i, n_j, P_i, P_j) P(S_{i,j} | n_i, n_j, P_i, P_j)} \tag{12}$$

Since we have assumed a uniform prior for $S_{i,j}$ (assumption 4), the prior terms vanish, leaving

$$P(R_{i,j}^t | E_i, E_j, P_i, P_j) = \frac{P(k | S_{i,j} = P_{min}, n_i, n_j, P_i, P_j)}{\sum_{k \le S_{i,j} \le P_{min}} P(k | n_i, n_j, P_i, P_j)} \tag{13}$$

The second step of the derivation is to find a suitable expression for

$$P(k | S_{i,j}, n_i, n_j, P_i, P_j)$$

The probability can be written out fully as:

$$P(k|S_{i,j}, n_i, n_j, P_i, P_j) = \frac{\displaystyle\sum_{\substack{E_i \subset U_i : |E_i| = n_i \\ E_j \subset U_j : |E_j| = n_j \\ |E_i \cap E_j| = k}} P(E_i, E_j | S_{i,j}, n_i, n_j, P_i, P_j)}{\displaystyle\sum_{\substack{E_i \subset U_i : |E_i| = n_i \\ E_j \subset U_j : |E_j| = n_j}} P(E_i, E_j | S_{i,j}, n_i, n_j, P_i, P_j)} \qquad (14)$$

By assumption 2, $P(E_i, E_j) = P(E_i)P(E_j)$. By Lemma 1, all $P(E_i)$ terms are equal, since they are all sets of size $n_i$, and likewise for $P(E_j)$ terms. Thus, to get the desired probability expression, we simply need to count the number of ways of taking subsets from the two urns such that they share $k$ properties.

$$P(k|S_{i,j}, n_i, n_j, P_i, P_j) = \frac{\displaystyle\sum_{\substack{E_i \subset U_i : |E_i| = n_i \\ E_j \subset U_j : |E_j| = n_j \\ |E_i \cap E_j| = k}} 1}{\displaystyle\sum_{\substack{E_i \subset U_i : |E_i| = n_i \\ E_j \subset U_j : |E_j| = n_j}} 1} \qquad (15)$$

$$= \frac{\mathrm{Count}(k, n_i, n_j | S_{i,j}, P_i, P_j)}{\mathrm{Count}(n_i, n_j | S_{i,j}, P_i, P_j)} \qquad (16)$$

There are $\binom{P_i}{n_i}$ ways of picking each set $E_i$, so

$$\mathrm{Count}(n_i, n_j | S_{i,j}, P_i, P_j) = \binom{P_i}{n_i}\binom{P_j}{n_j} \qquad (17)$$

To complete the derivation, we need an expression for $\mathrm{Count}(k, n_i, n_j | S_{i,j}, P_i, P_j)$. This involves splitting the relevant sets into several parts. First $U_i$ and $U_j$ each contain some shared and unshared properties. Let $T_{i,j} = U_i \cap U_j$, $V_i = U_i - T_{i,j}$, and $V_j = U_j - T_{i,j}$. Second, the selected sets from each urn, $E_i$ and $E_j$, each have properties that come from the set of shared properties and the set of unshared properties. Let $K = E_i \cap E_j$, $F_i = (E_i \cap T_{i,j}) - K$, and $F_j = (E_j \cap T_{i,j}) - K$.

With these sets defined, each set $E_i$ and $E_j$ is composed of three distinct subsets: the shared subset ($K$); a subset also selected from the shared potential properties, $T_{i,j}$, but which is not shared ($F_i$ and $F_j$); and the remaining elements, which are chosen from the complements of the shared properties ($V_i$ and $V_j$). Since the subsets are distinct, we can count them separately and multiply the results to arrive at the final count.

The number of ways of selecting the shared subset is clearly $\binom{S_{i,j}}{k}$. The sizes of $F_i$ and $F_j$ are unknown, however, so we must sum over all possibilities. Let $r = |F_i|$, and $s = |F_j|$. There are $S_{i,j} - k$ remaining shared potential properties in $T_{i,j}$ from which to choose the $r + s$ elements of $F_i$ and $F_j$, and then $\binom{r+s}{s}$ ways to split the two into distinct subsets. There are $n_i - (k + r)$ elements left to choose in $E_i$, and $n_j - (k + s)$ elements left to choose in $E_j$. These must be selected from the unshared potential properties in $V_i$ and $V_j$, which have sizes $P_i - S_{i,j}$ and $P_j - S_{i,j}$ respectively. Putting these pieces together, we have

$$\text{Count}(k, n_i, n_j | S_{i,j}, P_i, P_j) =$$

$$\binom{S_{i,j}}{k} \sum_{r,s} \binom{S_{i,j}-k}{r+s} \binom{r+s}{s} \binom{P_i - S_{i,j}}{n_i - (k+r)} \binom{P_j - S_{i,j}}{n_j - (k+s)} \tag{18}$$

The ranges for $r$ and $s$ are somewhat involved. They must obey the following constraints:

1. $r, s \geq 0$

2. $r \geq n_i - k - P_i + S_{i,j}$

3. $s \geq n_j - k - P_j + S_{i,j}$

4. $r \leq n_i - k$

5. $s \leq n_j - k$

6. $r + s \leq S_{i,j} - k$

Plugging Equation 18 into Equation 16, and that in turn into Equation 13 yields the desired result. □

## Appendix B. Fast Calculation of the Extracted Shared Property Model

The ESP model can be expensive to calculate if done the wrong way. We use two techniques to speed up the calculation immensely. For reference, the full formulation of the model is:

$$P(R_{i,j}^t | k, n_i, n_j, P_i, P_j) =$$

$$\frac{\binom{P_{min}}{k} \sum_{r,s \geq 0} \binom{S_{i,j}-k}{r+s} \binom{r+s}{r} \binom{P_i - P_{min}}{n_i - (k+r)} \binom{P_j - P_{min}}{n_j - (k+s)}}{\sum_{k \leq S_{i,j} \leq P_{min}} \binom{S_{i,j}}{k} \sum_{r,s \geq 0} \binom{S_{i,j}-k}{r+s} \binom{r+s}{r} \binom{P_i - S_{i,j}}{n_i - (k+r)} \binom{P_j - S_{i,j}}{n_j - (k+s)}} \tag{19}$$

Note that the equation involves three sums, ranging over $O(P_{min})$, $O(n_i)$, and $O(n_j)$ values respectively. In effect, this is $O(n^3)$ in the number of extractions for a string. Furthermore, each step requires the expensive operation of calculating binomial coefficients. Fortunately, there are several easy ways to drastically speed up this calculation.

First, Stirling's approximation can be used to calculate factorials (and therefore the binomial function). Stirling's approximation is given by:

$$n! \approx \sqrt{\pi \left(2n + \frac{1}{3}\right)} \left(\frac{n^n}{e^n}\right)$$

To avoid underflow and overflow errors, log probabilities are used everywhere possible. This calculation can then be done using a few simple multiplications and logarithm calculations. Stirling's formula converges to $n!$ like $O(\frac{1}{n})$; in practice it proved to be accurate enough of an approximation of $n!$ for $n > 100$. In ESP's implementation, all other values of $n!$ are calculated once, and stored for future use.

Second, the calculation of $P(k|n_1, n_2, P_1, P_2)$ can be sped up by simplifying the expression to get rid of two of the sums. The result is the following equivalent expression, assuming without loss of generality that $P_2 \leq P_1$:

$$P(k|n_1, n_2, P_1, P_2) = \frac{\binom{P_2+1}{n_2+1} \sum_{r=k}^{n_2} \binom{r}{k} \binom{P_1-r}{n_1-k}}{\binom{P_2}{n_2} \binom{P_1}{n_1}} \tag{20}$$

This simplification removes two of the sums, and therefore changes the complexity of calculating ESP from $O(P_2 n_2 n_1)$ to $O(n_2)$. This was sufficient for our data set, but on larger data sets it might be necessary to introduce sampling techniques to improve the efficiency even further.

## Appendix C. A Better Bound on the Number of Comparisons Made by the RESOLVER Clustering Algorithm

Section 4 showed that the RESOLVER clustering algorithm initially makes $O(N \log N)$ comparisons between strings in the data, where $N$ is the number of extractions. Heuristic methods like the Canopies method (McCallum et al., 2000) require $O(M^2)$ comparisons, where $M$ is the number of distinct strings in the data. We claim that $O(N \log N)$ is asymptotically better than $O(M^2)$ for Zipf-distributed data.

Zipf-distributed data is controlled by a shape parameter, which we call $z$. The claim above holds true for any shape parameter $z < 2$, as shown below. Fortunately, in natural data the shape parameter is usually very close to $z = 1$, and in RESOLVER data it was observed to be $z < 1$.

Let $S$ be the set of distinct strings in a set of extractions $D$. For each $s \in S$, let freq$(s)$ denote the number of times that $s$ appears in the extractions. Thus $|D| = \sum_{s \in S}$ freq$(s)$. Let $M = |S|$ and $N = |D|$.

**Proposition 3** *If $S$ has an observed Zipf distribution with shape parameter $z$, then*

1. *if $z < 1$, $N = \Theta(M)$*

2. *if $z = 1$, $N = \Theta(M \log M)$*

3. *if $z > 1$, $N = \Theta(M^z)$*

*Proof:* Let $s_1, \ldots, s_M$ be the elements of $S$ in rank order from highest frequency string ($s_1$) to lowest frequency string ($s_M$). Since $S$ has an observed Zipf distribution with shape parameter $z$, freq$(s_i) = \frac{M^z}{i^z}$. Given the assumptions, $z$ and $M$ determine the number of extractions made:

$$N_{M,z} = \sum_{s \in S} \text{freq}(s) \tag{21}$$

$$= \sum_{1 \leq i \leq M} \frac{M^z}{i^z} \tag{22}$$

We can build a recurrence relation for the value of $N$ as $M$ changes (holding $z$ constant) by noting that

$$N_{2M,z} = \sum_{1 \le i \le 2M} \frac{(2M)^z}{i^z} \tag{23}$$

$$= (2M)^z \sum_{1 \le i \le M} \frac{1}{i^z} + (2M)^z \sum_{M+1 \le i \le 2M} \frac{1}{i^z} \tag{24}$$

$$= 2^z N_{M,z} + f_z(M) \tag{25}$$

where $f_z(M) = \sum_{M+1 \le i \le 2M} \frac{(2M)^z}{i^z}$.

There are two important properties of $f_z(M)$.

1. Note that every term in the sum for $f_z(M)$ is less than $\frac{(2M)^z}{M^z} = 2^z$. Thus $f_z(M)$ is bounded above by $2^z \cdot M$, so if $z$ is held constant, $f_z(M) = O(M)$.

2. Every term in the sum is at least 1, so $f_z(M) \ge M$ and $f_z(M) = \Omega(M)$; combining these two facts yields $f_z(M) = \Theta(M)$.

These two properties of $f_z(M)$ will be used below.

We can now use the recurrence relation and the Master Recurrence Theorem (Cormen, Leiserson, & Rivest, 1990) to prove the three claims of the proposition. For reference, the Master Recurrence Theorem states the following:

**Theorem 1** *Let $a \ge 1$ and $b \ge 1$ be constants, let $f(n)$ be a function, and let $T(n)$ be defined on the non-negative integers by the recurrence*

$$T(n) = aT(n/b) + f(n)$$

*Then T(n) can be bounded asymptotically as follows.*

1. *If $f(n) = O(n^{\log_b a - \epsilon})$ for some constant $\epsilon > 0$, then $T(n) = \Theta(n^{\log_b a})$*

2. *If $f(n) = \Theta(n^{\log_b a})$, then $T(n) = \Theta(n^{\log_b a} \log n)$*

3. *If $f(n) = \Omega(n^{\log_b a + \epsilon})$, for some constant $\epsilon > 0$, and if $af(n/b) \le cf(n)$ for some constant $c < 1$ and all sufficiently large $n$, then $T(n) = \Theta(f(n))$.*

First consider the case where $z > 1$. The recurrence for $N_{M,z}$ can clearly be made to fit the form for Theorem 1 by setting $a = 2^z$, $b = 2$, and $f = f_z(M)$. Since $f_z(M)$ is bounded above by $2^z \cdot M = O(M)$, it is also clearly bounded above by $O(M^{\log_b a - \epsilon}) = O(M^{z-\epsilon})$, where we can take $\epsilon$ to be anything in $(0, z-1)$. Thus the case one of Theorem 1 applies, and $N_{M,z} = \Theta(M^{\log_b a}) = \Theta(M^z)$.

Next consider the case where $z = 1$. Since $f_{z=1}(M) = \Theta(M)$ and $\Theta(M^{\log_b a}) = \Theta(M)$, case two of Theorem 1 applies. Thus $N_{M,z=1} = \Theta(M^{\log_b a} \log M) = \Theta(M \log M)$.

Finally, consider the case where $z < 1$. Unfortunately, the regularity condition in case 3 of Theorem 1 does not hold for $f_z(M)$. Instead of using Theorem 1, we resort to a proof by induction.

Specifically, we show by induction that whenever $z < 1$ and $M \geq 2$, $N_{M,z} \leq c \cdot M$, where $c = \text{Max}(\frac{2^z + 1}{2}, \frac{2^z}{2 - 2^z})$. First, consider the case where $M = 2$:

$$
\begin{aligned}
N_{2,z} &= \sum_{i=1}^{2} \frac{2^z}{i^z} \\
&= 2^z + 1 \\
&= \frac{2^z + 1}{2} \cdot 2 \\
&\leq c \cdot M
\end{aligned}
$$

We now prove the induction case.

$$
\begin{aligned}
N_{M,z} &\leq 2^z N_{M/2,z} + f_z(M/2) \\
&\leq 2^z cM/2 + f_z(M/2) \quad \text{(by the induction hypothesis)} \\
&\leq 2^z cM/2 + 2^z M/2 \\
&= cM \cdot (2^{z-1} + 2^{z-1}/c) \\
&\leq cM \cdot (2^{z-1} + 2^{z-1} \frac{2 - 2^z}{2^z}) \quad \text{(by the definition of } c) \\
&= cM
\end{aligned}
$$

□

The data used in RESOLVER experiments in Section 4 had a shape parameter $z < 1$, so the bound on the number of comparisons made was $O(N \log N) = O(M \log M)$. For $z = 1$, the bound is $O(N \log N) = O(M \log M \log(M \log M)) = O(M \log^2 M)$. For $z > 1$, the bound is $O(M^z \log M)$. Not until $z = 2$ would the asymptotic performance of $O(M^2 \log M)$ have been worse than $O(M^2)$. If past experience is any guide, such a high value for $z$ is unlikely for extractions from naturally occurring text.